\documentclass[journal]{IEEEtran}
\usepackage{amsmath,amsfonts}
\usepackage{array}
\usepackage{algorithmic}
\usepackage[ruled,lined,linesnumbered,ruled]{algorithm2e}  
\usepackage{multirow}
\usepackage{graphicx} 
\usepackage{float} 
\usepackage[caption=false,font=footnotesize,labelfont=rm,textfont=rm]{subfig}
\usepackage{CJKutf8}
\usepackage{textcomp}
\usepackage{stfloats}
\usepackage{url}
\usepackage{verbatim}
\usepackage{graphicx}
\usepackage{cite}
\usepackage{lipsum}
\begin{document}
\begin{CJK}{UTF8}{gbsn}
\title{\textbf{Recurrent Stochastic Configuration Networks with Incremental Blocks}}

\author{
  Gang Dang \\
  State Key Laboratory of Synthetical Automation for Process Industries \\
  Northeastern University, Shenyang 110819, China\\
   Dianhui Wang 
  \thanks{\textit{\underline{Corresponding author}}: 
\textbf{dh.wang@deepscn.com}}\\
  State Key Laboratory of Synthetical Automation for Process Industries \\
  Northeastern University, Shenyang 110819, China \\
 Research Center for Stochastic Configuration Machines\\
  China University of Mining and Technology, Xuzhou 221116, China\\
}
\maketitle
\newtheorem{remark}{\bf Remark}           

\begin{abstract}                          
Recurrent stochastic configuration networks (RSCNs) have shown promise in modelling nonlinear dynamic systems with order uncertainty due to their advantages of easy implementation, less human intervention, and strong approximation capability. This paper develops the original RSCNs with block increments, termed block RSCNs (BRSCNs), to further enhance the learning capacity and efficiency of the network. BRSCNs can simultaneously add multiple reservoir nodes (subreservoirs) during the construction. Each subreservoir is configured with a unique structure in the light of a supervisory mechanism, ensuring the universal approximation property. The reservoir feedback matrix is appropriately scaled to guarantee the echo state property of the network. Furthermore, the output weights are updated online using a projection algorithm, and the persistent excitation conditions that facilitate parameter convergence are also established. Numerical results over a time series prediction, a nonlinear system identification task, and two industrial data predictive analyses demonstrate that the proposed BRSCN performs favourably in terms of modelling efficiency, learning, and generalization performance, highlighting their significant potential for coping with complex dynamics. 
\end{abstract}


\section{Introduction}
\IEEEPARstart{N}{owdays}, using neural networks (NNs) to analyze nonlinear dynamic systems has received considerable attention \cite{ref1,ref2,ref3}. As a class of data-driven techniques, the performance of NNs is significantly influenced by input variables. However, due to the uncertainties or changes in the controlled plant and external disturbances, the input order often varies over time, leading to systems with unknown dynamic orders. Recurrent neural networks (RNNs) have feedback connections between neurons, which can store the historical information and use them to handle the uncertainty caused by the selected input variables. Unfortunately, RNNs employ the error back-propagation (BP) algorithm to train the network, which suffers from the sensitivity of learning rate, slow convergence, and local minima \cite{ref4,ref5,ref511}. Reservoir computing (RC) provides an alternative scheme for training RNNs, which utilizes a large-scale sparsely connected reservoir to capture the state information and obtain an adaptable readout \cite{ref6,ref7}. As a class of randomized learning algorithms, RC effectively overcomes the limitations of gradient-based methods and encompasses various versions of RNNs such as echo state networks (ESNs) \cite{ref8} and liquid state machines (LSMs) \cite{ref9}.

In recent years, ESNs have been widely applied for tackling complex dynamics due to their computational simplicity, fast learning speed, and strong nonlinear processing capability \cite{ref10,ref11,ref12}. Distinguished from other RC approaches, ESNs have the echo state property (ESP), that is, the reservoir state $\mathbf{x}\left( n \right)$ is the echo of the input and $\mathbf{x}\left( n \right)$ should asymptotically depend only on the driving input signal \cite{ref8}. This unique characteristic makes them well-suited for temporal data analysis. However, ESNs face challenges in random parameter selection and structure setting, significantly affecting the model performance. To address the issues of structural design, researchers have introduced several approaches, including the simple circular reservoir (SCR) to minimize computational complexity while maintaining effectiveness \cite{ref13}, the leaky integrator ESN (LIESN) to enhance the model flexibility \cite{ref14}, deep ESNs to enrich the feature representation by constructing stacked structures \cite{ref15,ref16}, and pruning and growing strategies to adaptively adjust the reservoir topology \cite{ref17, ref18}. Nevertheless, properly setting the learning parameters for these methods is quite challenging in practical applications. While some optimization algorithms can aid in obtaining improved network parameters \cite{ref19,ref20,ref21}, they are constrained by complex iterations and their sensitivity to the initial state and learning rate. Furthermore, the aforementioned approaches cannot guarantee the model's universal approximation property, which is essential for data modelling theory and practice. According to \cite{ref22,ref23}, a randomized learner model exhibits excellent learning and generalization performance when it is incrementally built using a data-dependent random parameter scope, with the structural construction guided by certain theoretical principles. \vspace{-0.01cm}

In 2017, Wang and Li pioneered an innovative randomized learner model, termed stochastic configuration networks (SCNs) \cite{ref24}, which randomly assign the weights and biases in the light of a supervisory mechanism. Built on the SCN concept, in \cite{ref25}, we introduced a recurrent version of SCNs (RSCNs) to create a class of randomized universal approximators for temporal data. However, RSCNs add reservoir nodes using the point incremental strategy, which may result in numerous iterations during the construction process for large-scale applications. This paper develops RSCNs with block increments, termed block RSCNs (BRSCNs). BRSCNs start with a small-sized reservoir and stochastically configure subreservoirs based on the block recurrent stochastic configuration (BRSC) algorithm. Subsequently, the output weights are updated online using the projection algorithm \cite{ref26}. Experimental results demonstrate that the BRSCN outperforms other models in terms of learning and generalization performance, underscoring its effectiveness in modelling nonlinear dynamic systems. Therefore, the proposed approach offers several advantages.
\begin{itemize}
  \item [1)] 
 The block increments of reservoir nodes and batch assignment of random parameters effectively accelerate the construction process.      
  \item [2)]
The universal approximation property and echo state property of RSCNs are naturally inherited, enabling strong nonlinear processing and temporal data analysis capabilities.
  \item [3)]
The conditions for persistent excitation, which facilitate the dynamic adjustment of output weights based on the projection algorithm, are established to ensure the convergence of learning parameters.
  \item [4)]
The impact of subreservoir size on model performance is carefully considered, allowing BRSCNs to achieve sound performance for both learning and generalization by setting appropriate block sizes.
\end{itemize}

The remainder of this paper is organized as follows. Section II reviews the related knowledge of ESNs and RSCNs. Section III details BRSCNs with the algorithm description and relevant properties. Section IV  presents the parameter learning and convergence analysis. Section V reports the experimental results. Finally, Section VI concludes this paper.

\section{Related work}
In this section, two related works are introduced, including the well-known echo state networks and recurrent stochastic configuration networks.
\subsection{Echo state networks}
The ESN can be regarded as a simplified version of RNN, which utilizes a large-scale sparsely connected reservoir to transform input signals into a high-dimensional state space \cite{ref8}. The random parameters are generated from a fixed uniform distribution and remain constant during training. Only the output weights need to be calculated by the least square method. 

Given an ESN model,
\begin{equation}
\label{eq1}
{\bf{x}}(n) = g({{\bf{W}}_{{\rm{in}}}}{\bf{u}}(n) + {{\bf{W}}_{\mathop{\rm r}\nolimits} }{\bf{x}}(n - 1) + {\bf{b}}),
\end{equation}
\begin{equation}
\label{eq2}
{\bf{y}}(n) = {{\bf{W}}_{\rm out}}\left( {{\bf{x}}(n),{\bf{u}}(n)} \right),
\end{equation}
where $\mathbf{u}(n)\in {{\mathbb{R}}^{K}}$ is the input signal; $\mathbf{x}(n)\in {{\mathbb{R}}^{N}}$ is the internal state of the reservoir; ${{\mathbf{W}}_{\text{in}}}\in {{\mathbb{R}}^{N\times K}},{{\mathbf{W}}_{\rm r}}\in {{\mathbb{R}}^{N\times N}}$ represent the input and reservoir weights, respectively; $\mathbf{b}$ is the bias; ${{\mathbf{W}}_{\rm out}}\in {{\mathbb{R}}^{L\times \left( N+K \right)}}$ is the output weight; $K$ and $L$ are the dimensions of input and output; and $g$ is the activation function. ${{\mathbf{W}}_{\text{in}}},{{\mathbf{W}}_{\text{r}}},\mathbf{b}$ are generated from the uniform distribution $\left[ -\lambda ,\lambda  \right].$ The value of $\lambda $ has a significant impact on the model performance. The original ESNs use a fixed $\lambda $, which may lead to poor performance. Scholars have focused on optimizing the weight scope, and some promising results have been reported in \cite{ref20,ref21}. However, the optimization process inevitably increases the complexity of the algorithm. Therefore, selecting a data-dependent and adjustable $\lambda $ is crucial to improve the effectiveness and efficiency of the resulting model.

Define $\mathbf{X}\text{=}\left[ \left( \mathbf{x}(1),\mathbf{u}(1) \right),\ldots ,\left( \mathbf{x}({{n}_{\max }}),\mathbf{u}({{n}_{\max }}) \right) \right]$, where ${{n}_{\max }}$ is the number of training samples and the output is
\begin{equation}
\label{eq3}
{\bf{Y}} = \left[ {{\bf{y}}\left( 1 \right),{\bf{y}}\left( 2 \right),...,{\bf{y}}\left( {{n_{max}}} \right)} \right] = {\bf{W}}_{{\mathop{\rm out}\nolimits} }^{}{\bf{X}}.
\end{equation}
The output weight ${{\mathbf{W}}_{\rm out}}$ can be calculated by the least square method, that is,
\begin{equation}
\label{eq5}
{\bf{W}}_{{\mathop{\rm out}\nolimits} }^ \top  = {\left( {{\bf{X}}{{\bf{X}}^ \top }} \right)^{ - 1}}{\bf{X}}{{\bf{T}}^ \top },
\end{equation}
where $\mathbf{T}=\left[ \mathbf{t}\left( 1 \right),\mathbf{t}\left( 2 \right),...\mathbf{t}\left( {{n}_{max}} \right) \right]$ is the desired output. $\mathbf{x}(0)$ usually starts with a zero matrix, and a few warm-up samples are used to minimize the influence of the initial states. 

\subsection{Recurrent stochastic configuration networks}
\begin{figure}[htbp]
\vspace{-0.3cm}
	\begin{center}
		\includegraphics[width=9cm]{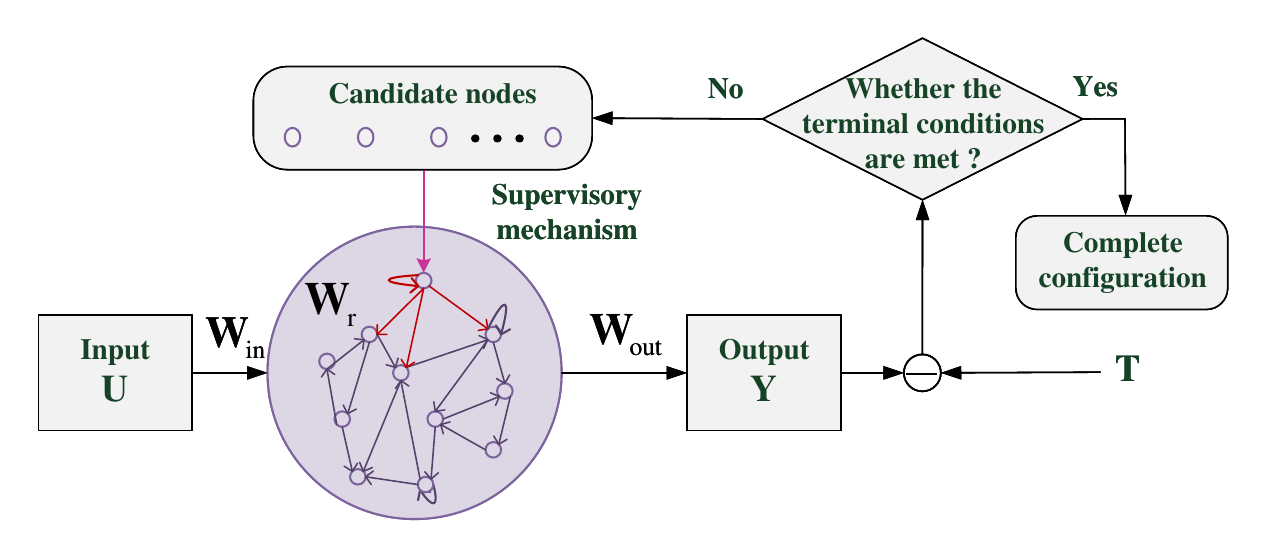}
		\caption{Architecture of the basic RSCN.}
		\label{fig1}
	\end{center}
\vspace{-0.3cm}
\end{figure}

This section reviews our proposed RSCNs \cite{ref25}, in which the random weights and biases are assigned in the light of a supervisory mechanism. This innovative learning scheme effectively addresses the issue of randomized neural network parameter selection and structure design, while theoretically ensuring the universal approximation performance of the network. With benefits such as high learning efficiency, less human intervention, and strong approximation ability, RSCNs have demonstrated promising potential for modelling complex dynamics. The architecture of the RSCN is shown in Fig. \ref{fig1}.

Considering an RSCN model constructed by Eq. (\ref{eq1}) and Eq. (\ref{eq2}), the weights and biases of the first reservoir node are randomly assigned, where ${{\mathbf{W}}_{\operatorname{in},1}}=\left[ \begin{matrix}
   w_{\operatorname{i}\operatorname{n}}^{1,1} & w_{\operatorname{i}\operatorname{n}}^{1,2} & \cdots  & w_{\operatorname{i}\operatorname{n}}^{1,K}  \\
\end{matrix} \right],$ ${{\mathbf{W}}_{\operatorname{r},1}}\text{=}w_{\operatorname{r}}^{1,1}$, ${{\mathbf{b}}_{1}}={{b}_{1}},$ $w_{\operatorname{in}}^{1,j},w_{\operatorname{r}}^{1,1},{{b}_{1}}\in \left[ -\lambda ,\lambda  \right]$. Given the input signals $\mathbf{U}\text{=}\left[ \mathbf{u}(1),\mathbf{u}(2),\ldots ,\mathbf{u}({{n}_{\max }}) \right]$, the residual error between the current model output and desired output is $e_N=\mathbf{Y}-\mathbf{T}$. When $e_N$ does not satisfy the preset error tolerance $\varepsilon$, it is necessary to generate a new random basis function ${{g}_{N+1}}$ based on the supervisory mechanism. As shown in Fig. \ref{fig1}, the incremental construction process of RSCNs can be summarized as follows.

Step 1: Initialize parameters: random weight sequence $\mathbf{\gamma }\text{=}\left\{ {{\lambda }_{\min }},{{\lambda }_{\min }}+\Delta \lambda ,...,{{\lambda }_{\max }} \right\}$, error threshold $\varepsilon$, the maximum number of stochastic configurations ${{G}_{\max }}$, current reservoir size $N=1$, the maximum reservoir size ${{N}_{\max }}$, and residual error ${{e}_{0}}={{e}_{N}}$.

Step 2: Assign $\left[ \begin{matrix}
   w_{\operatorname{i}\operatorname{n}}^{N+1,1} & w_{\operatorname{i}\operatorname{n}}^{N+1,2} & \cdots  & w_{\operatorname{i}\operatorname{n}}^{N+1,K}  \\
\end{matrix} \right],$ ${{\mathbf{W}}_{\mathop{\rm r},N+1}}$, and ${{b}_{N+1}}$ stochastically in ${{G}_{\max }}$ times from an adjustable uniform distribution $\left[ -{{\lambda }},{{\lambda }} \right]$. Construct the reservoir weight matrix with a special structure, where new node weights are assigned to the primary nodes and itself, while other nodes don’t have connections to the new node. The reservoir weights can be expressed as\vspace{-0.35cm}

\begin{small}
\begin{equation}
\label{eq6}
\begin{array}{l}
{{\bf{W}}_{\mathop{\rm r},2}}{\rm{ = }}\left[ {\begin{array}{*{20}{c}}
{w_{\mathop{\rm r}\nolimits} ^{1,1}}&0\\
{w_{\mathop{\rm r}\nolimits} ^{2,1}}&{w_{\mathop{\rm r}\nolimits} ^{2,2}}
\end{array}} \right],\\
{{\bf{W}}_{\mathop{\rm r},3}}{\rm{ = }}\left[ {\begin{array}{*{20}{c}}
{w_{\mathop{\rm r}\nolimits} ^{1,1}}&0&0\\
{w_{\mathop{\rm r}\nolimits} ^{2,1}}&{w_{\mathop{\rm r}\nolimits} ^{2,2}}&0\\
{w_{\mathop{\rm r}\nolimits} ^{3,1}}&{w_{\mathop{\rm r}\nolimits} ^{3,2}}&{w_{\mathop{\rm r}\nolimits} ^{3,3}}
\end{array}} \right],\\
 \ldots \\
{{\bf{W}}_{\mathop{\rm r},N{\rm{ + }}1}}{\rm{ = }}\left[ {\begin{array}{*{20}{c}}
{w_{\mathop{\rm r}\nolimits} ^{1,1}}&0& \cdots &0&0\\
{w_{\mathop{\rm r}\nolimits} ^{2,1}}&{w_{\mathop{\rm r}\nolimits} ^{2,2}}& \cdots &0&0\\
 \vdots & \vdots & \vdots & \vdots & \vdots \\
{w_{\mathop{\rm r}\nolimits} ^{N,1}}&{w_{\mathop{\rm r}\nolimits} ^{N,2}}& \cdots &{w_{\mathop{\rm r}\nolimits} ^{N,N}}&0\\
{w_{\mathop{\rm r}\nolimits} ^{N{\rm{ + 1,1}}}}&{w_{\mathop{\rm r}\nolimits} ^{N{\rm{ + 1,2}}}}& \cdots &{w_{\mathop{\rm r}\nolimits} ^{N{\rm{ + }}1,N}}&{w_{\mathop{\rm r}\nolimits} ^{N{\rm{ + }}1,N + 1}}
\end{array}} \right].
\end{array}
\end{equation}
\end{small}
The input weights and biases are defined as
\begin{equation} \label{eq7}
\begin{array}{l}
{{\bf{W}}_{{\rm{in}},2}}{\rm{ = }}\left[ {\begin{array}{*{20}{l}}
{w_{{\mathop{\rm i}\nolimits} {\mathop{\rm n}\nolimits} }^{1,1}}&{w_{{\mathop{\rm i}\nolimits} {\mathop{\rm n}\nolimits} }^{1,2}}& \cdots &{w_{{\mathop{\rm i}\nolimits} {\mathop{\rm n}\nolimits} }^{1,K}}\\
{w_{{\mathop{\rm i}\nolimits} {\mathop{\rm n}\nolimits} }^{2,1}}&{w_{{\mathop{\rm i}\nolimits} {\mathop{\rm n}\nolimits} }^{2,2}}& \cdots &{w_{{\mathop{\rm i}\nolimits} {\mathop{\rm n}\nolimits} }^{2,K}}
\end{array}} \right],\\
 \ldots \\
{{\bf{W}}_{{\rm{in}},N + 1}}{\rm{ = }}\left[ {\begin{array}{*{20}{c}}
{w_{{\mathop{\rm i}\nolimits} {\mathop{\rm n}\nolimits} }^{1,1}}&{w_{{\mathop{\rm i}\nolimits} {\mathop{\rm n}\nolimits} }^{1,2}}& \cdots &{w_{{\mathop{\rm i}\nolimits} {\mathop{\rm n}\nolimits} }^{1,K}}\\
{w_{{\mathop{\rm i}\nolimits} {\mathop{\rm n}\nolimits} }^{2,1}}&{w_{{\mathop{\rm i}\nolimits} {\mathop{\rm n}\nolimits} }^{2,2}}& \cdots &{w_{{\mathop{\rm i}\nolimits} {\mathop{\rm n}\nolimits} }^{2,K}}\\
 \vdots & \vdots & \vdots & \vdots \\
{w_{{\mathop{\rm i}\nolimits} {\mathop{\rm n}\nolimits} }^{N,1}}&{w_{{\mathop{\rm i}\nolimits} {\mathop{\rm n}\nolimits} }^{N,2}}& \cdots &{w_{{\mathop{\rm i}\nolimits} {\mathop{\rm n}\nolimits} }^{N,K}}\\
{w_{{\mathop{\rm i}\nolimits} {\mathop{\rm n}\nolimits} }^{N + 1,1}}&{w_{{\mathop{\rm i}\nolimits} {\mathop{\rm n}\nolimits} }^{N + 1,2}}& \cdots &{w_{{\mathop{\rm i}\nolimits} {\mathop{\rm n}\nolimits} }^{N + 1,K}}
\end{array}} \right],
\end{array}
\end{equation}
where ${{\mathbf{b}}_{2}}={{\left[ {{b}_{1}},{{b}_{2}} \right]}^{\top }},\ldots ,{{\mathbf{b}}_{N+1}}={{\left[ {{b}_{1}},\ldots {{b}_{N+1}} \right]}^{\top }}$.

Step 3: Seek the random basis function ${g_{N{\rm{ + }}1}}$ that satisfies the following inequality:
\begin{equation} \label{eq8}
{\left\langle {e_{N,q}^{},{g_{N{\rm{ + }}1}}} \right\rangle ^2} \ge b_g^2(1 - r - {\mu _{N + 1}})\left\| {{e_{N,q}}} \right\|_{}^2,q = 1,2,...L,
\end{equation}
where $0<r<1$, $\left\{ {{\mu }_{N+1}} \right\}$ is a non-negative real sequence satisfying $\underset{N\to \infty }{\mathop{\lim }}\,{{\mu }_{N+1}}=0$, and ${{\mu }_{N+1}}\le (1-r)$, $0<\left\| g \right\|<{{b}_{g}}$.

Step 4: Define a set of variables $\left[ {{\xi }_{N+1,1}},...,{{\xi }_{N+1,L}} \right]$ to select the node making the training error converge as soon as possible, that is,
\begin{equation} \label{eq9}
{\xi _{N + 1,q}} = \frac{{{{\left( {e_{N,q}^ \top {g_{N + 1}}} \right)}^2}}}{{g_{N + 1}^ \top {g_{N + 1}}}} - \left( {1 - {\mu _{N + 1}} - r} \right)e_{N,q}^ \top e_{N,q}^{}.
\end{equation}
The candidate node with the maximum ${{\xi }_{N+1}}=\sum\limits_{q=1}^{L}{{{\xi }_{N+1,q}}}$ is determined as the optimal adding node. 

Step 5: Evaluate the output weight by the global least square method:
\begin{equation} \label{eq10}
\begin{array}{l}
{\bf{W}}_{{\rm{out}},N{\rm{ + }}1}^{}{\rm{ = }}\left[ {{\bf{w}}_{{\rm{out}},1}^{},{\bf{w}}_{{\rm{out}},2}^{},...,{\bf{w}}_{{\rm{out}},N + 1 + K}^{}} \right]\\
{\kern 1pt} {\kern 1pt} {\kern 1pt} {\kern 1pt} {\kern 1pt}  {\kern 1pt} {\kern 1pt} {\kern 1pt} {\kern 1pt} {\kern 1pt} {\kern 1pt} {\kern 1pt} {\kern 1pt} {\kern 1pt} {\kern 1pt} {\kern 1pt} {\kern 1pt} {\kern 1pt} {\kern 1pt} {\kern 1pt} {\kern 1pt} {\kern 1pt} {\kern 1pt} {\kern 1pt} {\kern 1pt} {\kern 1pt} {\kern 1pt} {\kern 1pt} {\kern 1pt} {\kern 1pt} {\kern 1pt} {\kern 1pt} {\kern 1pt} {\kern 1pt} {\kern 1pt} {\kern 1pt} {\kern 1pt} {\kern 1pt} {\kern 1pt} {\kern 1pt}  = \mathop {\arg \min }\limits_{{\bf{W}}_{{\rm{out}}}^{}} \left\| {{\bf{T}} - {\bf{W}}_{{\rm{out}}}^{}{{\bf{X}}_{N + 1}}} \right\|_2^2,
\end{array}
\end{equation}
where \small ${{\mathbf{X}}_{N+1}}\text{=}\left[ \left( {{\mathbf{x}}_{N+1}}\left( 1 \right),\mathbf{u}\left( 1 \right) \right),\ldots ,\left( {{\mathbf{x}}_{N+1}}\left( {{n}_{\max }} \right),\mathbf{u}\left( {{n}_{\max }} \right) \right) \right]$. \normalsize

Step 6: Calculate the residual error ${{e}_{N+1}}$ and update ${{e}_{0}}:={{e}_{N+1}}$, $N=N+1$. Repeat steps 2-5 until ${{\left\| {{e}_{0}} \right\|}_{F}}<\varepsilon $ or $N\ge {{N}_{\max }}$.

Finally, we can obtain $\underset{N\to \infty }{\mathop{\lim }}\,{{\left\| {{e}_{N}} \right\|}_{F}}=0$. 

\begin{remark}
To mitigate the risk of overfitting, an additional condition is imposed to regulate the addition of hidden nodes. Moreover, a step size ${{N}_{\text{step}}}$ (${{N}_{\text{step}}}<N$) is utilized in the following early stopping criterion:
\begin{equation}
\label{eq101}
{\left\| {{e_{{\rm{val}},N - {N_{{\rm{step}}}}}}} \right\|_F} \le {\left\| {{e_{{\rm{val}},N - {N_{{\rm{step}}}} + 1}}} \right\|_F} \le  \ldots  \le {\left\| {{e_{{\rm{val}},N}}} \right\|_F},
\end{equation}
where ${{e}_{\text{val},N}}$ denotes the validation residual error with $N$ hidden nodes and ${{\left\| \bullet  \right\|}_{F}}$ represents the $F$ norm. If Eq. (\ref{eq101}) is satisfied, the number of hidden nodes will be adjusted to $N-{{N}_{\text{step}}}$.
\end{remark}

\section{Block incremental recurrent stochastic configuration networks}
This section details the proposed BRSCNs, including the algorithm description and proofs of the echo state property and the universal approximation property.
\begin{figure}[htbp]
	\begin{center}
		\includegraphics[width=9cm]{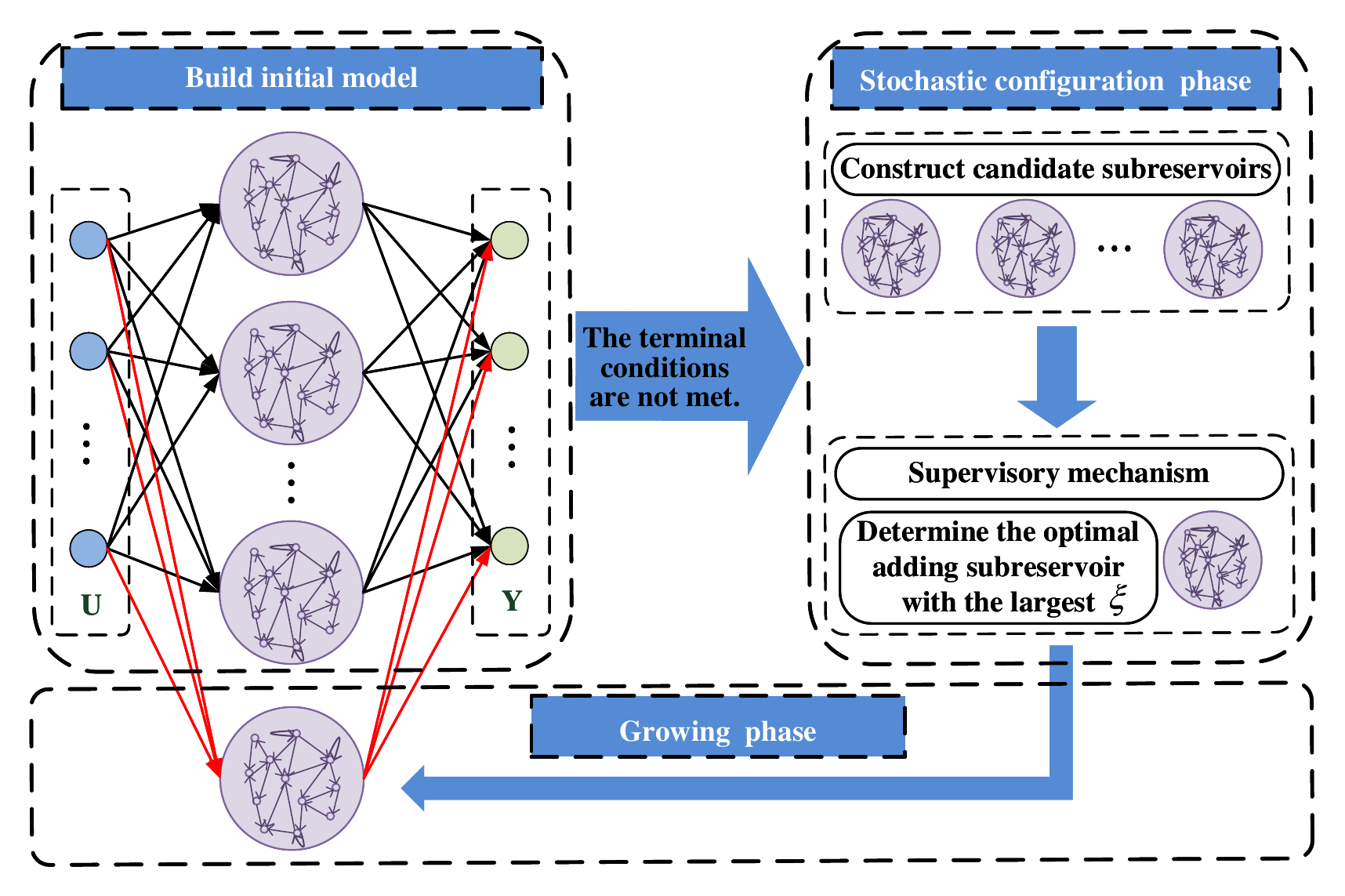}
		\caption{Architecture of the basic BRSCN.}
		\label{fig2}
	\end{center}
 \vspace{-0.8cm}
\end{figure}
\subsection{Algorithm description}
As shown in Fig. \ref{fig2}, the BRSCN adds reservoir nodes with block increments, and each block can be viewed as a subreservoir. Given the input $\mathbf{U}\text{=}\left[ \mathbf{u}(1),\ldots ,\mathbf{u}({{n}_{\max }}) \right]$ and desired output $\mathbf{T}=\left[ \mathbf{t}\left( 1 \right),...\mathbf{t}\left( {{n}_{max}} \right) \right]$, assume that the first subreservoir with $N$ nodes has been built, that is, 
\begin{equation} \label{eq11}
\begin{array}{*{20}{c}}
{{{\bf{x}}^{\left( 1 \right)}}(n) = g({\bf{W}}_{{\rm{in,}}N}^{\left( 1 \right)}{\bf{u}}(n) + {\bf{W}}_{{\rm{r,}}N}^{\left( 1 \right)}{{\bf{x}}^{\left( 1 \right)}}(n - 1) + {\bf{b}}_N^{\left( 1 \right)})}\\
{{\bf{Y}} = {\bf{W}}_{{\mathop{\rm out}\nolimits} }^{}{{\bf{X}}^{\left( 1 \right)}}}
\end{array},
\end{equation}
where $\mathbf{W}_{\text{in,}N}^{\left( 1 \right)}$, $\mathbf{W}_{\text{r,}N}^{\left( 1 \right)}$, $\mathbf{b}_{N}^{\left( 1 \right)}$, and ${{\mathbf{x}}^{\left( 1 \right)}}(n)$ are the input weight, reservoir weight, bias, and reservoir state of the first subreservoir, respectively. $\mathbf{W}_{\operatorname{out}}^{{}}$ is the output weight, and ${{\mathbf{X}}^{\left( 1 \right)}}\text{=}\left[ {{\mathbf{x}}^{\left( 1 \right)}}\left( 1 \right),{{\mathbf{x}}^{\left( 1 \right)}}\left( 2 \right),\ldots ,{{\mathbf{x}}^{\left( 1 \right)}}\left( {{n}_{\max }} \right) \right]$.

Let $j=1$ and calculate the current error ${{e}_{0}}:={{e}_{j}}=\mathbf{Y}-\mathbf{T}$. If ${{\left\| e_{0}^{{}} \right\|}_{F}}>\varepsilon $, we need to add subreservoirs under the supervisory mechanism until satisfying termination conditions. Assign $\mathbf{W}_{\text{in,}N}^{\left( j+1 \right)}$, $\mathbf{W}_{\text{r,}N}^{\left( j+1 \right)}$ and $\mathbf{b}_{N}^{\left( j+1 \right)}$ stochastically in ${{G}_{\max }}$ times from an adjustable uniform distribution $\left[ -\lambda ,\lambda  \right]$ to obtain the candidates of the subreservoir state $\mathbf{X}_{{}}^{\left( j+1 \right)1},\mathbf{X}_{{}}^{\left( j+1 \right)2},\ldots ,\mathbf{X}_{{}}^{\left( j+1 \right),{{G}_{\max }}}$. Substitute $\mathbf{X}_{{}}^{\left( j+1 \right)1},\mathbf{X}_{{}}^{\left( j+1 \right)2},\ldots ,\mathbf{X}_{{}}^{\left( j+1 \right),{{G}_{\max }}}$ into the following inequality constraint:
\begin{equation} \label{eq12}
\begin{array}{l}
(1 - r - {\mu _{j + 1}})\left\| {e_{j,q}^{}} \right\|_2^2 - \frac{{{{\left\langle {e_{j,q}^{},{{\bf{X}}^{\left( {j + 1} \right),i}}} \right\rangle }^2}}}{{\left\langle {{{\bf{X}}^{\left( {j + 1} \right),i}},{{\bf{X}}^{\left( {j + 1} \right),i}}} \right\rangle }} \le 0,\\
q = 1,2, \ldots ,L,{\kern 1pt} {\kern 1pt} {\kern 1pt} {\kern 1pt} {\kern 1pt} {\kern 1pt} {\kern 1pt} i = 1,2, \ldots ,{G_{\max }},
\end{array}
\end{equation}
where $\left\{ {{\mu }_{j+1}} \right\}$ is a non-negative real sequence satisfies $\underset{j\to \infty }{\mathop{\lim }}\,{{\mu }_{j\text{+}1}}=0$ and ${{\mu }_{j\text{+}1}}\le \left( 1-r \right)$.

\begin{remark}
To ensure the echo state property of the model, the reservoir weight $\mathbf{W}_{\text{r,}N}^{\left( j+1 \right)}$ needs to be scaled by
\begin{equation} \label{eq13}
{\bf{W}}_{{\rm{r,}}N}^{\left( {j + 1} \right)} \leftarrow \frac{\alpha }{{\rho  _{\max }^{j + 1}}}{\bf{W}}_{{\rm{r,}}N}^{\left( {j + 1} \right)},
\end{equation}
where $0<\alpha<1$ is the scaling factor, and $\rho _{\max }^{j+1}$ is the maximum eigenvalue value of $\mathbf{W}_{\text{r,}N}^{\left( j+1 \right)}$.
\end{remark}

Seek the subreservoirs that satisfy Eq. (\ref{eq12}) and define a set of variables ${{\xi }_{j+1}}\text{=}\left[ {{\xi }_{j+1,1}},{{\xi }_{j+1,2}},...,{{\xi }_{j+1,L}} \right]$,
\begin{equation} \label{eq14}
{\xi _{j{\rm{ + }}1,q}} = \frac{{{{\left\langle {e_{j,q}^{},{{\bf{X}}^{\left( {j + 1} \right)}}} \right\rangle }^2}}}{{\left\langle {{{\bf{X}}^{\left( {j + 1} \right)}},{{\bf{X}}^{\left( {j + 1} \right)}}} \right\rangle }} - (1 - r - {\mu _{j + 1}})e_{j,q}^ \top e_{j,q}^{}.
\end{equation}
A larger positive value of $\xi _{j+1}^{{}}\text{=}\sum\limits_{q=1}^{L}{{{\xi }_{j+1,q}}}$ implies a better configuration of the adding subreservoir.

Calculate the current training and validation residual error $e_{j+1}^{{}}$ and $e_{{\rm{val}},j + 1}^{}$, and a step size ${{j}_{\text{step}}}$ (${{j}_{\text{step}}}<j$) is used in the early stopping criterion, that is,
\begin{equation}
\label{eq141}
{\left\| {{e_{{\rm{val}},j - {j_{{\rm{step}}}}}}} \right\|_F} \le {\left\| {{e_{{\rm{val}},j - {j_{{\rm{step}}}} + 1}}} \right\|_F} \le  \ldots  \le {\left\| {{e_{{\rm{val}},j}}} \right\|_F}.
\end{equation}
Renew ${{e}_{0}}:=e_{j+1}^{{}}$, $j=j+1$, and continue to add subreservoirs until ${{\left\| e_{0}^{{}} \right\|}_{F}}\le\varepsilon $ or $j \ge J_{\max }^{}$ or Eq. (\ref{eq141}) is met. The general construction process can be summarized as follows:\vspace{-0.3cm}

\begin{scriptsize}
\begin{equation} \label{eq17}
\begin{array}{*{20}{c}}
\begin{array}{l}
\left[ {\begin{array}{*{20}{c}}
{{{\bf{x}}^{\left( 1 \right)}}(n)}\\
{{{\bf{x}}^{\left( 2 \right)}}(n)}\\
 \vdots \\
{{{\bf{x}}^{\left( j \right)}}(n)}
\end{array}} \right] = g\left( {\left[ {\begin{array}{*{20}{c}}
{{\bf{W}}_{{\rm{in,}}N}^{\left( 1 \right)}}\\
{{\bf{W}}_{{\rm{in,}}N}^{\left( 2 \right)}}\\
 \vdots \\
{{\bf{W}}_{{\rm{in,}}N}^{\left( j \right)}}
\end{array}} \right]{\bf{u}}(n) + } \right.\\
\left. {\left[ {\begin{array}{*{20}{c}}
{{\bf{W}}_{{\rm{r,}}N}^{\left( 1 \right)}}&0&0&0\\
0&{{\bf{W}}_{{\rm{r,}}N}^{\left( 2 \right)}}&0&0\\
 \vdots & \vdots & \ddots & \vdots \\
0&0&0&{{\bf{W}}_{{\rm{r,}}N}^{\left( j \right)}}
\end{array}} \right]\left[ {\begin{array}{*{20}{c}}
{{{\bf{x}}^{\left( 1 \right)}}(n - 1)}\\
{{{\bf{x}}^{\left( 2 \right)}}(n - 1)}\\
 \vdots \\
{{{\bf{x}}^{\left( j \right)}}(n - 1)}
\end{array}} \right] + \left[ {\begin{array}{*{20}{c}}
{{\bf{b}}_N^{\left( 1 \right)}}\\
{{\bf{b}}_N^{\left( 2 \right)}}\\
 \vdots \\
{{\bf{b}}_N^{\left( j \right)}}
\end{array}} \right]} \right)
\end{array}\\
{{\bf{Y}} = {\bf{W}}_{{\mathop{\rm out}\nolimits} }^{}\left[ {\begin{array}{*{20}{c}}
{{{\bf{X}}^{\left( 1 \right)}}}&{{{\bf{X}}^{\left( 2 \right)}}}& \ldots &{{{\bf{X}}^{\left( j \right)}}}
\end{array}} \right]}\\
{\left[ {{\bf{W}}_{{\mathop{\rm out}\nolimits} }^{\left( 1 \right)*},{\bf{W}}_{{\mathop{\rm out}\nolimits} }^{\left( 2 \right)*}, \ldots ,{\bf{W}}_{{\mathop{\rm out}\nolimits} }^{\left( j \right)*}} \right] = \mathop {\arg \min }\limits_{{\bf{W}}_{{\mathop{\rm out}\nolimits} }^{}} \left\| {{\bf{T}} - \sum\limits_{k = 1}^j {{\bf{W}}_{{\mathop{\rm out}\nolimits} }^{\left( k \right)}{{\bf{X}}^{\left( k \right)}}} } \right\|},
\end{array}
\end{equation}
\end{scriptsize}where $\mathbf{W}_{\operatorname{out}}^{\left( j \right)}$ is the output weight corresponding to $j$-th subreservoir.

Finally, we have $\underset{j\to \infty }{\mathop{\lim }}\,\left\| \mathbf{T}-F_{j}^{{}} \right\|=0$, where $F_{j}^{{}}$ is the final output with $j$ subreservoirs. A complete algorithm description of the proposed BRSCNs is summarized in Algorithm 1.

\begin{algorithm}[t]\footnotesize
    \caption{DeepRSC}\label{algo1}	
    \KwIn{Training inputs $\mathbf{U}\text{=}\left[ \mathbf{u}(1),\ldots ,\mathbf{u}({{n}_{\max }}) \right]$, training outputs $\mathbf{T}=\left[ \mathbf{t}\left( 1 \right),...\mathbf{t}\left( {{n}_{max}} \right) \right]$, validation inputs ${{\bf{U}}_{{\rm{val}}}}{\rm{ = }}\left[ {{{\bf{u}}_{{\rm{val}}}}(1),{{\bf{u}}_{{\rm{val}}}}(2), \ldots } \right]$, validation outputs ${{\bf{T}}_{{\rm{val}}}}{\rm{ = }}\left[ {{{\bf{t}}_{{\rm{val}}}}(1),{{\bf{t}}_{{\rm{val}}}}(2), \ldots } \right]$, an initial subreservoir size $N$, a maximum number of subreservoirs $J_{\max }^{{}}$, a step size ${{J}_{\text{step}}}$, a training error threshold $\varepsilon $, the positive scalars $\mathbf{\gamma }\text{=}\left\{ {{\lambda }_{1}},{{\lambda }_{2}},...,{{\lambda }_{\max }} \right\}$, and the maximum number of stochastic configurations ${{G}_{\max }}$.}
    \KwOut{BRSCN}
    Randomly assign $\mathbf{W}_{\text{in,}N}^{\left( 1 \right)}$, $\mathbf{W}_{\text{r,}N}^{\left( 1 \right)}$, and $\mathbf{b}_{N}^{\left( 1 \right)}$ according to the sparsity of the reservoir from $\left[ -\lambda ,\lambda  \right]$. Calculate the model output $\mathbf{Y}$ and current error ${{e}_{j}}$ and  ${{e}_{\text{val},j}}$, where $j=1$. Set the initial residual error ${{e}_{0}}:={{e}_{j}}$, $0<r<1$, $\mathbf{\Omega },\mathbf{D}:=\left[ {\kern 1pt}{\kern 1pt} \right]$;\\
    \While {$j<{{J}_{\max }}$ AND ${{\left\| {{e}_{0}} \right\|}_{F}}>\varepsilon $}{
    \If{Eq. (\ref{eq141}) is not met}{
        \For{$\lambda \in \mathbf{\gamma }$,}{
            \For{$l=1,2,\ldots ,{{G}_{\max }}$,}{
                Randomly assign $\mathbf{W}_{\text{in,}N}^{\left( j+1 \right)}$, $\mathbf{W}_{\text{r,}N}^{\left( j+1 \right)}$, and $\mathbf{b}_{N}^{\left( j+1 \right)}$ from $\left[ -\lambda ,\lambda  \right]$;\\
                Calculate the subreservoir state $\mathbf{X}_{{}}^{\left( j+1 \right)}$;\\
                Set ${{\mu }_{j\text{+}1}}=\frac{1-r}{\left( j+1 \right)*N}$ and calculate ${{\xi }_{j\text{+}1,q}}$ based on Eq. (\ref{eq14});\\
                \If{$\min \left\{ {{\xi }_{j+\text{1,1}}},{{\xi }_{j\text{+}1,2}},...,{{\xi }_{j\text{+1},L}} \right\}\ge 0$}{
                    Save $\mathbf{W}_{\text{in,}N}^{\left( j+1 \right)}$, $\mathbf{W}_{\text{r,}N}^{\left( j+1 \right)}$, and $\mathbf{b}_{N}^{\left( j+1 \right)}$ in $\mathbf{D}$, and $\xi _{j+1}^{{}}\text{=}\sum\limits_{q=1}^{L}{{{\xi }_{j+1,q}}}$ in $\mathbf{\Omega }$;\\
                    \Else{Go back to \textbf{Step 5}}\
                }            
            }
            \If{$\mathbf{D}$ is not empty}{
                Find $\mathbf{W}_{\text{in,}N}^{\left( j+1 \right)*}$, $\mathbf{W}_{\text{r,}N}^{\left( j+1 \right)*}$, and $\mathbf{b}_{N}^{\left( j+1 \right)*}$ that maximize ${{\xi }_{j\text{+}1}}$ in $\mathbf{\Omega }$, and get $\mathbf{X}_{{}}^{\left( j+1 \right)*}$;\\
                 \textbf{Break} (go to \textbf{Step 26});\\    
                \Else{Randomly take $\tau \in \left( 0,1-r \right)$, update $r=r+\tau $, and return \textbf{Step 5};}\
                }         
        }
        Obtain $\mathbf{X}_{{}}^{\left( 1 \right)*}$, $\mathbf{X}_{{}}^{\left( 2 \right)*}$,...,$\mathbf{X}_{{}}^{\left( j+1 \right)*}$ and calculate $\mathbf{W}_{\operatorname{out}}^{*}=\left[ \mathbf{W}_{\operatorname{out}}^{\left( 1 \right)*},\mathbf{W}_{\operatorname{out}}^{\left( 2 \right)*},\ldots ,\mathbf{W}_{\operatorname{out}}^{\left( j+1 \right)*} \right]$;\\
        Calculate $e_{j\text{+}1}^{{}}\text{=}e_{j}^{{}}-\mathbf{W}_{\operatorname{out}}^{\left( j+1 \right)*}\mathbf{X}_{{}}^{\left( j+1 \right)*}$;\\
        Update ${{e}_{0}}:={{e}_{j+1}}$, and $j=j+1$;\\
        \Else{Set the optimal reservoir size to $j:=j-{{j}_{\text{step}}}$, and obtain the current network parameters;\\
        \textbf{Break} (go to \textbf{Step 36});}\
        }
    }
    \textbf{Return} $\mathbf{W}_{\operatorname{out}}^{*}$, $\mathbf{W}_{\text{in,}N}^{\left( 1 \right)*},\mathbf{W}_{\text{in,}N}^{\left( 2 \right)*},\ldots ,\mathbf{W}_{\text{in,}N}^{\left( j \right)*}$, $\mathbf{W}_{\text{r,}N}^{\left( 1 \right)*},\mathbf{W}_{\text{r,}N}^{\left( 2 \right)*},\ldots ,\mathbf{W}_{\text{r,}N}^{\left( j \right)*}$, and $\mathbf{b}_{N}^{\left( 1 \right)*},\mathbf{b}_{N}^{\left( 2 \right)*},\ldots ,\mathbf{b}_{N}^{\left( j \right)*}$.
\end{algorithm}
\subsection{The echo state property for BRSCN}
One key distinguishing feature of ESN compared to other RC approaches is its echo state property. This unique property ensures that as the input sequence length approaches infinity, the discrepancy between two reservoir states driven by the same input sequence but with different initial conditions becomes negligible. The reliance on the initial state ${\bf{x}}\left( 0 \right)$ diminishes over time. Considering an ESN without output feedback, if the maximum singular value of the reservoir weight matrix is less than 1, the resulting learner model exhibits the echo state property \cite{ref8}.

\textbf{Theorem 1.} Given a BRSCN with $J$ subreservoirs, the maximum singular values of each subreservoir weight matrix are $\sigma _{\max }^{1},\ldots ,\sigma _{\max }^{J}$. If the scaling factor in Eq. (\ref{eq13}) is selected as 
\begin{equation} \label{eq181}
0 < \alpha  < \frac{{\rho _{\max }^j\left( {{\bf{W}}_{\text{r},N}^{\left( j \right)})} \right)}}{{\sigma _{\max }^j\left( {{\bf{W}}_{\text{r},N}^{\left( j \right)}} \right)}},
\end{equation}the built model holds the echo state property.

\textbf{Proof.} According to Eq. (\ref{eq17}), the reservoir weight matrix can be written as:

\begin{small}
\begin{equation} \label{eq18}
\begin{array}{l}
{\bf{W}}_{\rm{r}}^{} = \left[ {\begin{array}{*{20}{c}}
{{\bf{W}}_{{\rm{r,}}N}^{\left( 1 \right)}}&0&0&0\\
0&{{\bf{W}}_{{\rm{r,}}N}^{\left( 2 \right)}}&0&0\\
 \vdots & \vdots & \ddots & \vdots \\
0&0&0&{{\bf{W}}_{{\rm{r,}}N}^{\left( J \right)}}
\end{array}} \right]\\
 = diag\left\{ {{\bf{W}}_{{\rm{r,}}N}^{\left( 1 \right)},{\bf{W}}_{{\rm{r,}}N}^{\left( 2 \right)}, \ldots ,{\bf{W}}_{{\rm{r,}}N}^{\left( J \right)}} \right\}\\
 = diag\left\{ {{{\bf{P}}_1}{{\bf{S}}_1}{{\bf{Q}}_1}, \ldots ,{{\bf{P}}_J}{{\bf{S}}_J}{{\bf{Q}}_J}} \right\} = {\bf{P'}}diag\left\{ {{{\bf{S}}_1}, \ldots ,{{\bf{S}}_J}} \right\}{\bf{Q'}}\\
 = {\bf{P'}}diag\left\{ {\sigma _1^1,\sigma _2^1, \ldots ,\sigma _N^1, \ldots ,\sigma _1^J,\sigma _2^J, \ldots ,\sigma _N^J} \right\}{\bf{Q'}},
\end{array}
\end{equation}
\end{small}where ${{\mathbf{P}}_{j}}$ and ${{\mathbf{Q}}_{j}}$ are orthogonal matrices generated by SVD decomposition, $\mathbf{{P}'}=diag\left\{ {{\mathbf{P}}_{1}},\ldots ,{{\mathbf{P}}_{J}} \right\}$, $\mathbf{{Q}'}=diag\left\{ {{\mathbf{Q}}_{1}},\ldots ,{{\mathbf{Q}}_{J}} \right\}$, and ${{\mathbf{S}}_{j}}=diag\left\{ \sigma _{1}^{j},\sigma _{2}^{j},\ldots ,\sigma _{N}^{j} \right\}$ is composed of the singular values of the $j$-th subreservoir with $N$ nodes. Thus, it can be inferred that $diag\left\{ {{\mathbf{S}}_{1}},\ldots ,{{\mathbf{S}}_{J}} \right\}$ has the same singular values with $\mathbf{W}_{\text{r}}^{{}}$. Observe that
\begin{small}
 \begin{equation} \label{eq180}
\begin{array}{l}
\sigma _{\max }^j\left( {\frac{\alpha }{{\rho _{\max }^j\left( {{\bf{W}}_{\text{r},N}^{\left( j \right)}} \right)}}{\bf{W}}_{\text{r},N}^{\left( j \right)}} \right) = \frac{\alpha }{{\rho _{\max }^j\left( {{\bf{W}}_{\text{r},N}^{\left( j \right)}} \right)}}\sigma _{\max }^j\left( {{\bf{W}}_{\text{r},N}^{\left( j \right)}} \right)\\
 < \frac{{\rho _{\max }^j\left( {{\bf{W}}_{\text{r},N}^{\left( j \right)}} \right)}}{{\sigma _{\max }^j\left( {{\bf{W}}_{\text{r},N}^{\left( j \right)}} \right)}} \times \frac{1}{{\rho _{\max }^j\left( {{\bf{W}}_{\text{r},N}^{\left( j \right)}} \right)}}\sigma _{\max }^j\left( {{\bf{W}}_{\text{r},N}^{\left( j \right)}} \right) = 1.
\end{array}
\end{equation}   
\end{small}Then, we can easily obtain the maximum singular value of $\mathbf{W}_{\text{r}}^{{}}$ is less than 1, which completes the proof.
\vspace{-0.2cm}
\subsection{The universal approximation property for BRSCN}
BRSCNs generate subreservoirs in the light of the supervisory mechanism, enabling the network to effectively approximate any nonlinear mappings. This subsection presents the theoretical result of its universal approximation property, which is an extension proposed in \cite{ref27}.

Given a cost function
\begin{small}
\begin{equation} \label{eq19}
\begin{array}{l}
{J_{{\bf{W}}_{{\mathop{\rm out}\nolimits} }^{}}} = {\left\| {{\bf{T}} - F_{j + 1}^{}} \right\|^2}\\
 = {\left\| {{\bf{T}} - F_j^{} - {\bf{W}}_{{\mathop{\rm out}\nolimits} }^{\left( {j + 1} \right)}{{\bf{X}}^{\left( {j + 1} \right)}}} \right\|^2}\\
 = {\left\| {e_j^{} - {\bf{W}}_{{\mathop{\rm out}\nolimits} }^{\left( {j + 1} \right)}{{\bf{X}}^{\left( {j + 1} \right)}}} \right\|^2}\\
 = {\sum\limits_{q = 1}^L {\left( {e_{j,q}^{} - {\bf{W}}_{{\mathop{\rm out}\nolimits} ,q}^{\left( {j + 1} \right)}{{\bf{X}}^{\left( {j + 1} \right)}}} \right)} ^ \top }\left( {e_{j,q}^{} - {\bf{W}}_{{\mathop{\rm out}\nolimits} ,q}^{\left( {j + 1} \right)}{{\bf{X}}^{\left( {j + 1} \right)}}} \right)\\
 = \sum\limits_{q = 1}^L {\left( {{{\left\| {e_{j,q}^{}} \right\|}^2} - 2e_{j,q}^ \top \left( {{\bf{W}}_{{\mathop{\rm out}\nolimits} ,q}^{\left( {j + 1} \right)}{{\bf{X}}^{\left( {j + 1} \right)}}} \right)} \right.} \\
\left. { + \left( {{\bf{W}}_{{\mathop{\rm out}\nolimits} ,q}^{\left( {j + 1} \right)}{{\bf{X}}^{\left( {j + 1} \right)}}} \right){{\left( {{\bf{W}}_{{\mathop{\rm out}\nolimits} ,q}^{\left( {j + 1} \right)}{{\bf{X}}^{\left( {j + 1} \right)}}} \right)}^ \top }} \right)\\
 = {\left\| {e_j^{}} \right\|^2} - \sum\limits_{q{\rm{ = }}1}^L {\left( {2e_{j,q}^ \top \left( {{\bf{W}}_{{\mathop{\rm out}\nolimits} ,q}^{\left( {j + 1} \right)}{{\bf{X}}^{\left( {j + 1} \right)}}} \right)} \right.} \\
\left. { - \left( {{\bf{W}}_{{\mathop{\rm out}\nolimits} ,q}^{\left( {j + 1} \right)}{{\bf{X}}^{\left( {j + 1} \right)}}} \right){{\left( {{\bf{W}}_{{\mathop{\rm out}\nolimits} ,q}^{\left( {j + 1} \right)}{{\bf{X}}^{\left( {j + 1} \right)}}} \right)}^ \top }} \right),
\end{array}
\end{equation}    
\end{small}taking the derivative of Eq. (\ref{eq19}) with respect to $\mathbf{W}_{\operatorname{out},q}^{\left( j+1 \right)}$, yields
\begin{equation} \label{eq20}
\frac{{\partial {J_{{\bf{W}}_{{\mathop{\rm out}\nolimits} }^{}}}}}{{\partial {\bf{W}}_{{\mathop{\rm out}\nolimits} ,q}^{\left( {j + 1} \right)}}} =  - 2e_{j,q}^{}{{\bf{X}}^{\left( {j + 1} \right)}}^ \top  + 2{\bf{W}}_{{\mathop{\rm out}\nolimits} ,q}^{\left( {j + 1} \right)}{{\bf{X}}^{\left( {j + 1} \right)}}{{\bf{X}}^{\left( {j + 1} \right)}}^ \top .
\end{equation}
Then, we have
\begin{equation} \label{eq21}
{\bf{W}}_{{\mathop{\rm out}\nolimits} ,q}^{\left( {j + 1} \right)} = e_{j,q}^{}{{\bf{X}}^{\left( {j + 1} \right)}}^ \top {\left( {{{\bf{X}}^{\left( {j + 1} \right)}}{{\bf{X}}^{\left( {j + 1} \right)}}^ \top } \right)^{ - 1}}.
\end{equation}

\textbf{Theorem 2.} Assume span($\Gamma$) is dense on ${{L}_{2}}$ space. Given $0<r<1$ and a non-negative real sequence $\left\{ {{\mu }_{j+1}} \right\}$ satisfies $\underset{j\to \infty }{\mathop{\lim }}\,{{\mu }_{j+1}}=0$ and ${{\mu }_{j+1}}\le \left( 1-r \right)$. For $j=1,2...$, and $q=1,2,...,L$, define
\begin{equation} \label{eq22}
\begin{array}{l}
\delta _{j{\rm{ + }}1,q}^* = (1 - r - {\mu _{j{\rm{ + }}1}})\left\| {e_{j,q}^*} \right\|_{}^2,\\
\delta _{j{\rm{ + }}1}^* = \sum\limits_{q = 1}^L {\delta _{j{\rm{ + }}1,q}^*}.
\end{array}
\end{equation}
If the subreservoir state ${{\mathbf{X}}^{\left( j+1 \right)}}$ is generated by the following inequality constraint:
\begin{equation} \label{eq23}
\left\langle {e_j^ * ,{\bf{W}}_{{\mathop{\rm out}\nolimits} }^{\left( {j + 1} \right)}{{\bf{X}}^{\left( {j + 1} \right)}}} \right\rangle  \ge \delta _{j{\rm{ + }}1}^*,
\end{equation}
and the output weights are evaluated by Eq. (\ref{eq17}), we have $\underset{j\to \infty }{\mathop{\lim }}\,\left\| \mathbf{T}-F_{j+1}^{{}} \right\|=0$.\\

\textbf{Proof.} With simple computation, we have\vspace{-0.3cm}

\begin{small}
\begin{equation} \label{eq25}
\begin{array}{l}
\left\| {e_{j{\rm{ + }}1}^*} \right\|_2^2 - \left( {r + {\mu _{j{\rm{ + }}1}}} \right)\left\| {e_j^*} \right\|_2^2\\
= \sum\limits_{q = 1}^L {\left( {\left\langle {e_{j,q}^* - {\bf{W}}_{{\mathop{\rm out}\nolimits} ,q}^{\left( {j + 1} \right)}{{\bf{X}}^{\left( {j + 1} \right)}},e_{j,q}^* - {\bf{W}}_{{\mathop{\rm out}\nolimits} ,q}^{\left( {j + 1} \right)}{{\bf{X}}^{\left( {j + 1} \right)}}} \right\rangle } \right.} \\
 = \left. { - \left( {r + {\mu _{j{\rm{ + }}1}}} \right)\left\langle {e_{j,q}^*,e_{j,q}^*} \right\rangle } \right)\\
 = \sum\limits_{q = 1}^L {\left( {\left( {1 - r - {\mu _{j{\rm{ + }}1}}} \right)\left\langle {e_{j,q}^*,e_{j,q}^*} \right\rangle  - 2\left\langle {e_{j,q}^*,{\bf{W}}_{{\mathop{\rm out}\nolimits} ,q}^{\left( {j + 1} \right)}{{\bf{X}}^{\left( {j + 1} \right)}}} \right\rangle } \right.} \\
 = \left. { + \left\langle {{\bf{W}}_{{\mathop{\rm out}\nolimits} ,q}^{\left( {j + 1} \right)}{{\bf{X}}^{\left( {j + 1} \right)}},{\bf{W}}_{{\mathop{\rm out}\nolimits} ,q}^{\left( {j + 1} \right)}{{\bf{X}}^{\left( {j + 1} \right)}}} \right\rangle } \right)\\
 = \sum\limits_{q = 1}^L {\left( {\delta _{j{\rm{ + }}1,q}^* - e_{j,q}^ * {{\bf{X}}^{\left( {j + 1} \right)}}^ \top {{\left( {{{\bf{X}}^{\left( {j + 1} \right)}}{{\bf{X}}^{\left( {j + 1} \right)}}^ \top } \right)}^{ - 1}}{{\bf{X}}^{\left( {j + 1} \right)}}e_{j,q}^{ *  \top }} \right)} \\
 = \sum\limits_{q = 1}^L {\left( {\delta _{j{\rm{ + }}1,q}^* - {\bf{W}}_{{\mathop{\rm out}\nolimits} ,q}^{\left( {j + 1} \right)}{{\bf{X}}^{\left( {j + 1} \right)}}e_{j,q}^{ *  \top }} \right)} \\
 = \sum\limits_{q = 1}^L {\left( {\delta _{j{\rm{ + }}1,q}^* - \left\langle {e_{j,q}^*,{\bf{W}}_{{\mathop{\rm out}\nolimits} ,q}^{\left( {j + 1} \right)}{{\bf{X}}^{\left( {j + 1} \right)}}} \right\rangle } \right)} \\
 = \delta _{j{\rm{ + }}1}^* - \left\langle {e_j^*,{\bf{W}}_{{\mathop{\rm out}\nolimits} }^{\left( {j + 1} \right)}{{\bf{X}}^{\left( {j + 1} \right)}}} \right\rangle  \le 0.
\end{array}
\end{equation}
\end{small}Therefore, $\left\| e_{j\text{+}1}^{*} \right\|_{{}}^{2}\le \left( r+{{\mu }_{j\text{+}1}} \right)\left\| e_{j}^{*} \right\|_{{}}^{2}$ and we can easily establish 
\begin{equation} \label{eq26}
\left\| {e_{j{\rm{ + }}1}^*} \right\|_{}^2 \le r\left\| {e_j^*} \right\|_{}^2 + {\mu _{j{\rm{ + }}1}}\left\| {e_j^*} \right\|_{}^2.
\end{equation}Obviously, $\underset{j\to \infty }{\mathop{\lim }}\,{{\mu }_{j\text{+}1}}\left\| e_{j}^{*} \right\|_{{}}^{2}=0$. Combining Eq. (\ref{eq26}) and $0<r<1$, we can further obtain $\underset{j\to \infty }{\mathop{\lim }}\,\left\| e_{j\text{+}1}^{*} \right\|_{{}}^{2}=0$, indicating  $\underset{j\to \infty }{\mathop{\lim }}\,\left\| e_{j}^{*} \right\|=0$, which completes the proof.

\section{Parameters learning}
In this section, an online parameter learning strategy based on the projection algorithm is provided. Moreover, we investigate the persistent excitation conditions that facilitate parameter convergence and present the corresponding theoretical results. 
\subsection{Online learning of parameters}
Due to the dynamic changes in the actual industrial process, it is essential to update the model parameters timely to achieve a superior modelling performance using the updated model. Given a BRSCN model constructed by Eq. (\ref{eq17}) and denoted by ${\bf{g}}(n) = \left[ {{{\bf{x}}^{\left( 1 \right)}}(n), \ldots ,{{\bf{x}}^{\left( j \right)}}(n)} \right]$, a projection algorithm \cite{ref26} is applied to update the output weights ${\bf{W}}_{{\mathop{\rm out}\nolimits} }$.

Let ${\rm{H}} =\left\{ \mathbf{W}_{\operatorname{out}}^{{}}:\mathbf{y}(n)=\mathbf{W}_{\operatorname{out}}^{{}}\mathbf{g}(n) \right\}$ and define $\mathbf{W}_{\operatorname{out}}^{{}}(n-1)$ as the output weight at the $\left( n-1 \right)$-th step. Choose the closest weight to $\mathbf{W}_{\operatorname{out}}^{{}}(n-1)$, and find $\mathbf{W}_{\operatorname{out}}^{{}}(n)$ through minimizing the following cost function:
\begin{equation} \label{eq201}
\begin{array}{*{20}{c}}
{J = \frac{1}{2}{{\left\| {{\bf{W}}_{{\mathop{\rm out}\nolimits} }^{}\left( n \right) - {\bf{W}}_{{\mathop{\rm out}\nolimits} }^{}\left( {n - 1} \right)} \right\|}^2}}\\
{s.t.{\kern 1pt} {\kern 1pt} {\kern 1pt} {\kern 1pt} {\kern 1pt} {\kern 1pt} {\kern 1pt} {\kern 1pt} {\kern 1pt} {\kern 1pt} {\bf{y}}\left( n \right) = {\bf{W}}_{{\mathop{\rm out}\nolimits} }^{}\left( n \right){\bf{g}}(n)}
\end{array}.
\end{equation}
By introducing the Lagrange operator ${{\lambda }_{\text{p}}}$, we have
\begin{equation} \label{eq202}
\begin{array}{l}
{J_e} = \frac{1}{2}{\left\| {{\bf{W}}_{{\mathop{\rm out}\nolimits} }^{}\left( n \right) - {\bf{W}}_{{\mathop{\rm out}\nolimits} }^{}\left( {n - 1} \right)} \right\|^2}\\
{\kern 1pt} {\kern 1pt} {\kern 1pt} {\kern 1pt} {\kern 1pt} {\kern 1pt} {\kern 1pt} {\kern 1pt} {\kern 1pt} {\kern 1pt} {\kern 1pt} {\kern 1pt} {\kern 1pt} {\kern 1pt} {\kern 1pt} {\kern 1pt} {\kern 1pt} {\kern 1pt} {\kern 1pt} {\kern 1pt} {\kern 1pt} {\kern 1pt}  + {\lambda _{\rm{p}}}\left( {{\bf{y}}\left( n \right) - {\bf{W}}_{{\mathop{\rm out}\nolimits} }^{}\left( n \right){\bf{g}}(n)} \right).
\end{array}
\end{equation}
The necessary conditions for ${{J}_{e}}$ to be minimal are
\begin{equation} \label{eq203}
\frac{{\partial {J_e}}}{{\partial {\bf{W}}_{{\mathop{\rm out}\nolimits} }^{}\left( n \right)}} = 0{\kern 1pt} {\kern 1pt} {\kern 1pt} {\kern 1pt} {\kern 1pt} {\kern 1pt} {\kern 1pt} and{\kern 1pt} {\kern 1pt} {\kern 1pt} {\kern 1pt} {\kern 1pt} {\kern 1pt} \frac{{\partial {J_e}}}{{\partial {\lambda _{\rm{p}}}}} = 0{\kern 1pt}.
\end{equation}
Thus, we can obtain
\begin{equation} \label{eq204}
\left\{ {\begin{array}{*{20}{c}}
{{\bf{W}}_{{\mathop{\rm out}\nolimits} }^{}\left( n \right) - {\bf{W}}_{{\mathop{\rm out}\nolimits} }^{}\left( {n - 1} \right) - {\lambda _{\rm{p}}}{\bf{g}}{{(n)}^ \top } = 0}\\
{{\bf{y}}\left( n \right) - {\bf{W}}_{{\mathop{\rm out}\nolimits} }^{}\left( n \right){\bf{g}}(n) = 0}
\end{array}} \right.,
\end{equation}
\begin{equation} \label{eq205}
{\lambda _{\rm{p}}} = \frac{{{\bf{y}}\left( n \right) - {\bf{W}}_{{\mathop{\rm out}\nolimits} }^{}\left( {n - 1} \right){\bf{g}}(n)}}{{{\bf{g}}{{(n)}^ \top }{\bf{g}}(n)}}.
\end{equation}
Substituting Eq. (\ref{eq205}) into Eq. (\ref{eq204}), the online update rule for the output weights can be expressed as
\begin{equation} \label{eq206}
\begin{array}{l}
{\bf{W}}_{{\mathop{\rm out}\nolimits} }^{}(n) = {\bf{W}}_{{\mathop{\rm out}\nolimits} }^{}(n - 1)\\
{\kern 1pt}  + \frac{{{\bf{g}}{{(n)}^ \top }}}{{{\bf{g}}{{(n)}^ \top }{\bf{g}}(n)}}\left( {{\bf{y}}\left( n \right) - {\bf{W}}_{{\mathop{\rm out}\nolimits} }^{}\left( {n - 1} \right){\bf{g}}(n)} \right).
\end{array}
\end{equation}
To prevent division by zero, a small constant $c$ is added to the denominator. Furthermore, a coefficient $\gamma >0$ can be multiplied by the numerator to obtain the improved projection algorithm, that is,
\begin{equation} \label{eq207}
\begin{array}{l}
{\bf{W}}_{{\mathop{\rm out}\nolimits} }^{}(n) = {\bf{W}}_{{\mathop{\rm out}\nolimits} }^{}(n - 1)\\
{\kern 1pt}  + \frac{{\gamma {\bf{g}}{{(n)}^ \top }}}{{c + {\bf{g}}{{(n)}^ \top }{\bf{g}}(n)}}\left( {{\bf{y}}\left( n \right) - {\bf{W}}_{{\mathop{\rm out}\nolimits} }^{}\left( {n - 1} \right){\bf{g}}(n)} \right).
\end{array}
\end{equation}
\begin{remark}
    The detailed stability and convergence analysis of the proposed approach based on the projection algorithm has been presented in our previous work \cite{ref25}, and an enhanced condition is introduced to further improve the model’s stability. These theoretical results are crucial for evaluating whether the algorithm can achieve a stable solution over time, which directly influences the prediction performance and reliability of the model during the identification process.
\end{remark}

\subsection{Persistent excitation condition for parameter convergence}
This subsection offers a persistent excitation condition for online learning based on the projection algorithm. It necessitates that the input sequence exhibits sufficient richness or diversity over an extended time window, facilitating a comprehensive understanding of each state of the system. This theoretical result is essential to ensure that the parameters converge during the update process, ultimately resulting in accurate estimates.\\

\textbf{Theorem 3.} The convergence of parameters can be guaranteed if the input signal satisfies the following persistent excitation conditions:
\begin{equation} \label{eq208}
{\eta _1} \ge \int_{{n_0}}^{{n_0} + {n_{\rm{w}}}} {{\bf{g}}{{(n)}^ \top }{\bf{g}}(n)dn \ge {\eta _2}} ,
\end{equation}
\begin{equation} \label{eq209}
\Delta {P^{ - 1}}(n) \le 2\gamma {\eta _2} - {\gamma ^2}\eta _1^2,
\end{equation}
where $\eta $ is a positive constant, ${{n}_{\text{w}}}$ is the length of the time window, $\Delta {{P}^{-1}}\left( n \right)={{P}^{-1}}\left( n \right)-{{P}^{-1}}\left( n-1 \right)$, and $P\left( n \right)=\frac{1}{c+\mathbf{g}{{(n)}^{\top }}\mathbf{g}(n)}$.\\

\textbf{Proof.} The parameter estimation error can be calculated by
\begin{equation} \label{eq210}
{\bf{E}}\left( {n - 1} \right) = {{\bf{W}}_0} - {\bf{W}}_{{\mathop{\rm out}\nolimits} }^{}(n - 1),
\end{equation}
where $\mathbf{W}_{0}^{{}}$ is an ideal output weight satisfying $\mathbf{y}\left( n \right)\approx \mathbf{W}_{0}^{{}}\mathbf{g}(n)$. Combining Eq. (\ref{eq207}) and Eq. (\ref{eq210}), we have
\begin{equation} \label{eq211}
\begin{array}{l}
{\bf{E}}\left( n \right) = {{\bf{W}}_0} - {\bf{W}}_{{\mathop{\rm out}\nolimits} }^{}(n)\\
 = {{\bf{W}}_0} - {\bf{W}}_{{\mathop{\rm out}\nolimits} }^{}(n - 1)\\
 - \frac{{\gamma {\bf{g}}{{(n)}^ \top }}}{{c + {\bf{g}}{{(n)}^ \top }{\bf{g}}(n)}}\left( {{\bf{y}}\left( n \right) - {\bf{W}}_{{\mathop{\rm out}\nolimits} }^{}\left( {n - 1} \right){\bf{g}}(n)} \right)\\
 = {{\bf{W}}_0} - {\bf{W}}_{{\mathop{\rm out}\nolimits} }^{}(n - 1)\\
 - P(n)\gamma {\bf{g}}{(n)^ \top }\left( {{\bf{W}}_0^{}{\bf{g}}(n) - {\bf{W}}_{{\mathop{\rm out}\nolimits} }^{}\left( {n - 1} \right){\bf{g}}(n)} \right)\\
 = \left( {{\bf{I}} - P(n)\gamma {\bf{g}}{{(n)}^ \top }{\bf{g}}(n)} \right){\bf{E}}\left( {n - 1} \right).
\end{array}
\end{equation}
Define a Lyapunov function candidate, that is,
\begin{equation} \label{eq212}
V\left( n \right) = {\bf{E}}{\left( n \right)^ \top }{P^{ - 1}}\left( n \right){\bf{E}}\left( n \right).
\end{equation}
The change of $V\left( n \right)$ is denoted by $\Delta V\left( n \right)=V\left( n \right)-V\left( n-1 \right)$. Combining Eq. (\ref{eq211}) and Eq. (\ref{eq212}), we can obtain
\begin{equation} \label{eq213}
\begin{array}{l}
\Delta V\left( n \right)\\
 = {\bf{E}}{\left( n \right)^ \top }{P^{ - 1}}\left( n \right){\bf{E}}\left( n \right) - V\left( {n - 1} \right)\\
 = {\left( {\left( {{\bf{I}} - P(n)\gamma {\bf{g}}{{(n)}^ \top }{\bf{g}}(n)} \right){\bf{E}}\left( {n - 1} \right)} \right)^ \top }{P^{ - 1}}\left( n \right)\\
{\kern 1pt} {\kern 1pt} {\kern 1pt} {\kern 1pt} {\kern 1pt} {\kern 1pt} {\kern 1pt} {\kern 1pt} {\kern 1pt} {\kern 1pt} \left( {\left( {{\bf{I}} - P(n)\gamma {\bf{g}}{{(n)}^ \top }{\bf{g}}(n)} \right){\bf{E}}\left( {n - 1} \right)} \right)\\
{\kern 1pt} {\kern 1pt} {\kern 1pt} {\kern 1pt} {\kern 1pt} {\kern 1pt} {\kern 1pt} {\kern 1pt} {\kern 1pt}  - {\bf{E}}{\left( {n - 1} \right)^ \top }{P^{ - 1}}\left( {n - 1} \right){\bf{E}}\left( {n - 1} \right)\\
 = {\bf{E}}{\left( {n - 1} \right)^ \top }\left( {{P^{ - 1}}\left( n \right) - {P^{ - 1}}\left( {n - 1} \right) - 2\gamma {\bf{g}}{{(n)}^ \top }{\bf{g}}(n)} \right.\\
{\kern 1pt} {\kern 1pt} {\kern 1pt} {\kern 1pt} {\kern 1pt} {\kern 1pt} {\kern 1pt} {\kern 1pt} {\kern 1pt} \left. { + {\gamma ^2}P(n){\bf{g}}{{(n)}^ \top }{\bf{g}}(n){P^{ - 1}}\left( n \right){\bf{g}}{{(n)}^ \top }{\bf{g}}(n)} \right){\bf{E}}\left( {n - 1} \right).
\end{array}
\end{equation}
From Eq. (\ref{eq208}), it can be shown that ${{\eta }_{1}}\ge \mathbf{g}{{(n)}^{\top }}\mathbf{g}(n)\ge {{\eta }_{2}}$ and we have
\begin{equation} \label{eq214}
\begin{array}{l}
\Delta V\left( n \right) \le {\bf{E}}{\left( {n - 1} \right)^ \top }\left( {{P^{ - 1}}\left( n \right) - {P^{ - 1}}\left( {n - 1} \right)} \right.\\
{\kern 1pt} {\kern 1pt} {\kern 1pt} {\kern 1pt} {\kern 1pt} {\kern 1pt} {\kern 1pt} {\kern 1pt} {\kern 1pt} {\kern 1pt} {\kern 1pt} {\kern 1pt} {\kern 1pt} {\kern 1pt} {\kern 1pt} {\kern 1pt} {\kern 1pt} {\kern 1pt} {\kern 1pt} {\kern 1pt} {\kern 1pt} {\kern 1pt} {\kern 1pt} {\kern 1pt} {\kern 1pt} {\kern 1pt} {\kern 1pt} {\kern 1pt} {\kern 1pt} {\kern 1pt} {\kern 1pt} {\kern 1pt} {\kern 1pt} {\kern 1pt} {\kern 1pt} {\kern 1pt} {\kern 1pt} {\kern 1pt} {\kern 1pt} {\kern 1pt} {\kern 1pt} {\kern 1pt} {\kern 1pt} {\kern 1pt} {\kern 1pt} {\kern 1pt} {\kern 1pt} {\kern 1pt} {\kern 1pt} {\kern 1pt} {\kern 1pt} {\kern 1pt} {\kern 1pt} {\kern 1pt} {\kern 1pt} {\kern 1pt} {\kern 1pt} {\kern 1pt} {\kern 1pt} {\kern 1pt}  - 2\gamma {\eta _2}\left. { + {\gamma ^2}\eta _1^2} \right){\bf{E}}\left( {n - 1} \right).
\end{array}
\end{equation}
As Eq. (\ref{eq209}) holds, we can easily obtain $\Delta V\left( n \right)\le 0$. This demonstrates that updating the output weights based on the projection algorithm can guarantee the asymptotic convergence of the parameters, which completes the proof. 

\section{Experiment results}
In this section, the effectiveness of the BRSCN is tested on the Mackey-Glass time series prediction, a nonlinear system identification task, and two industrial data predictive analyses. The performance of the proposed BRSCN is compared with the original ESN and RSCN, and two block incremental ESNs, that is, growing ESN (GESN) \cite{ref18} and decoupled ESN (DESN) \cite{ref28}. The output weights are updated online through the projection algorithm \cite{ref26}, in response to the dynamic variations within the system. The normalized root means square error (NRMSE) is used to evaluate the model performance, that is,
\begin{equation} \label{eq28}
NRMSE=\sqrt{\frac{\sum\limits_{n=1}^{{{n}_{max}}}{{{\left( \mathbf{y}\left( n \right)-\mathbf{t}\left( n \right) \right)}^{2}}}}{{{n}_{max}}\operatorname{var}\left( \mathbf{t} \right)}},
\end{equation}
where $\operatorname{var}\left( \mathbf{t} \right)$ denotes the variance of the desired output.

The key parameters are taken as follows: the sparsity of the reservoir weight is set to $\left[ 0.01,0.03 \right]$, and the scaling factor of spectral radius $\alpha$ ranges from 0.5 to 1. For the ESN, DESN, and GESN, the scope setting of input and reservoir weights are set with $\lambda \in \left[ 0.1,1 \right]$. Specifically, for the DESN and GESN, the subreservoir size is set to 10. For RSC-based frameworks, the following parameters are taken as weight scale sequence $\left\{ 0.5,1,5,10,30,50,100 \right\}$, contractive sequence $r=\left[ 0.9,0.99,0.999,0.9999,0.99999 \right]$, the maximum number of stochastic configurations ${{G}_{\max }}=100$, and training tolerance ${{\varepsilon }}={{10}^{-6}}$. The initial reservoir size of RSCN is set to 5. The grid search method is used to identify the hyperparameters, including the reservoir size $N$ and the block size. Each experiment consists of 50 independent trials conducted under the same conditions, and the mean and standard deviation of NRMSE and training time are exploited to evaluate the model performance.
\begin{table*}[htbp]
\scriptsize
\caption{Performance comparison of different models on MG tasks.} \label{tb1}
\centering
\begin{tabular}{cccccc}
\hline
Datasets             & Models & Reservoir size & Training time            & Training NRMSE           & Testing NRMSE            \\ \hline
\multirow{5}{*}{MG}  & ESN    & 96             & 0.14271±0.03564          & 0.01172±0.00208          & 0.02573±0.00596          \\
                     & DESN    & 80             & \textbf{0.11035±0.05237} & 0.00634±0.00056          & 0.01722±0.00739          \\
                     & GESN   & 90             & 0.28762±0.19001          & 0.00405±0.00062          & 0.01563±0.00851          \\
                     & RSCN   & 68             & 0.95326±0.22615          & 0.00338±0.00045          & 0.01207±0.00619          \\
                     & BRSCN  & 50             & 0.76889±0.14271          & \textbf{0.00316±0.00047} & \textbf{0.01119±0.00142} \\ \hline
\multirow{5}{*}{MG1} & ESN    & 124            & 0.16983±0.05084          & 0.01572±0.00679          & 0.03983±0.01130          \\
                     & DESN    & 100            & \textbf{0.12938±0.07361} & 0.01192±0.00102          & 0.02736±0.00572          \\
                     & GESN   & 100            & 0.40426±0.15372          & 0.00673±0.00054          & 0.02218±0.00376          \\
                     & RSCN   & 79             & 0.88756±0.53245          & 0.00509±0.00047          & 0.01433±0.00378          \\
                     & BRSCN  & 70             & 0.83928±0.57240          & \textbf{0.00483±0.00069} & \textbf{0.01346±0.00221} \\ \hline
\multirow{5}{*}{MG2} & ESN    & 135            & 0.15339±0.07362          & 0.03091±0.01927          & 0.08309±0.01122          \\
                     & DESN    & 120            & \textbf{0.12982±0.04551} & 0.01647±0.00887          & 0.07028±0.00983          \\
                     & GESN   & 110            & 0.33823±0.22938          & 0.01009±0.00763          & 0.05128±0.00436          \\
                     & RSCN   & 105            & 1.21287±0.29201          & \textbf{0.00712±0.00550} & 0.03516±0.00249          \\
                     & BRSCN  & 110            & 1.15741±0.28523          & 0.00728±0.00115          & \textbf{0.03129±0.00285} \\ \hline
\end{tabular}
\vspace{-0.5cm}
\end{table*}
\begin{figure*}
\vspace{-0.3cm}
	\centering
	\subfloat{\includegraphics[width=7cm]{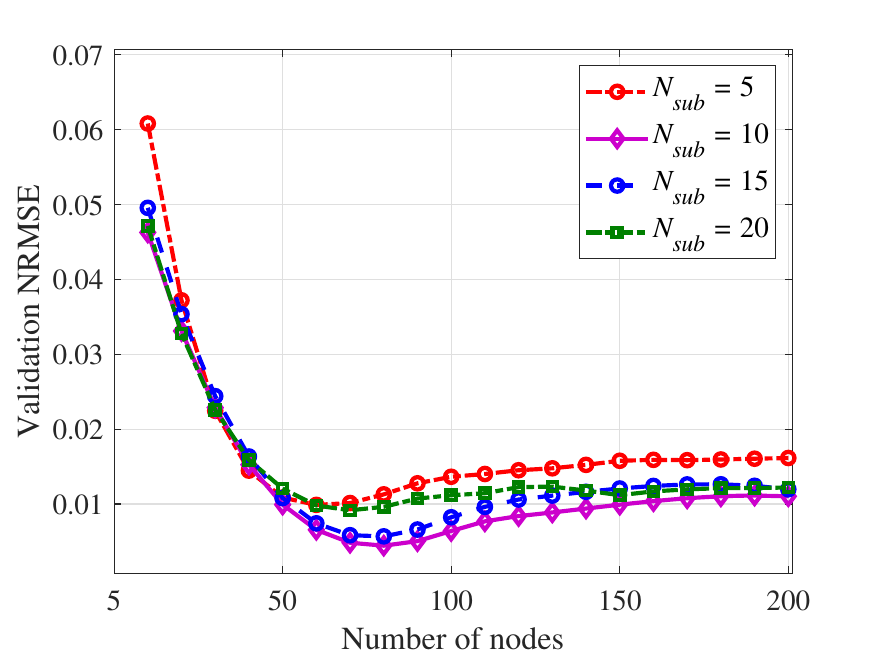}}
	\subfloat{\includegraphics[width=7cm]{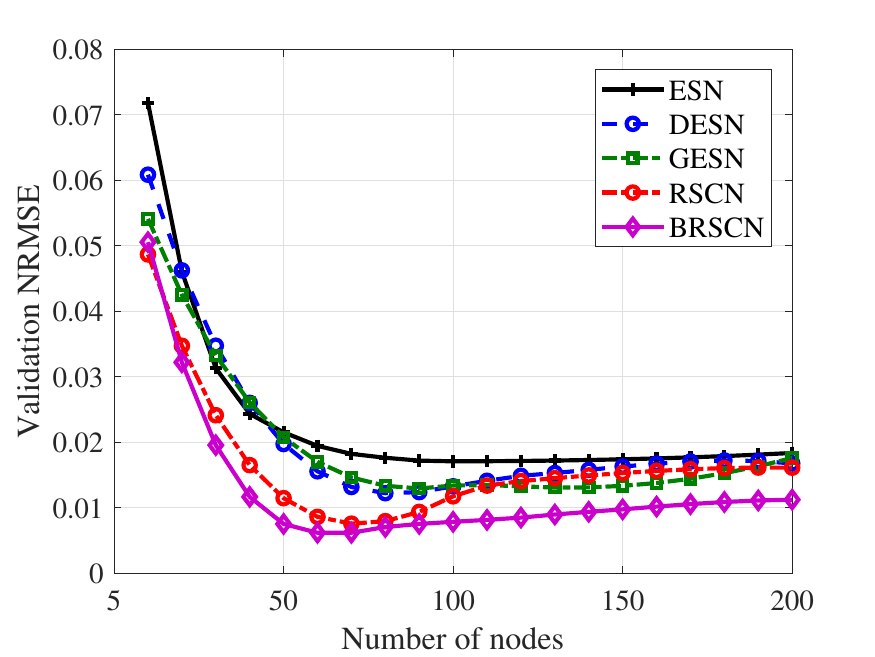}}
	\caption{Validation performance of different models on the MG task.}
	\label{fig3}
\vspace{-0.5cm}
\end{figure*}
\subsection{Mackey–Glass system (MGS)}
MGS is a classical chaotic system, which has been widely used for time series prediction and nonlinear system identification. A standard MGS can be described by the differential equations with time delays, that is,
\begin{equation} \label{eq29}
\frac{{du}}{{dn}} = \upsilon u\left( n \right) + \frac{{\alpha u\left( {n - \tau } \right)}}{{1 + u{{\left( {n - \tau } \right)}^{10}}}}.
\end{equation}
When $\tau >16.8$, the system transitions into a chaotic, aperiodic, non-convergent, and divergent state. According to \cite{ref8}, the parameters in Eq. (\ref{eq29}) are set to The initial values $\left\{ {y\left( 0 \right), \ldots ,y\left( \tau  \right)} \right\}$ are selected from $\left[ {0.1,1.3} \right]$. The inputs consist of $\left\{ {y\left( n \right),y\left( {n - 6} \right),y\left( {n - 12} \right),y\left( {n - 18} \right)} \right\}$, which are used to predict $y\left( {n + 6} \right)$. The second-order Runge-Kutta method is employed to generate 1177 sequence points. In our simulation, samples from time steps 1 to 500 are utilized for training the network, samples from 501 to 800 are used for validation, and the remaining samples serve for testing. The first 20 samples from each set are washed out. Moreover, considering the order uncertainty, we assume that certain orders are unknown and design two experimental setups. In the MG1 task, we select ${{\mathbf{u}}}\left( n \right)={{\left[ y\left( n-6  \right),y\left( n-12  \right),y\left( n-18  \right) \right]}}$ to predict $y\left( {n + 6} \right)$. In the MG2 task, the input is set as ${{\mathbf{u}}}\left( n \right)={{\left[ y\left( n-12  \right),y\left( n-18  \right) \right]}^{\top }}$. These settings are intentionally designed to assess the performance of the RSCN under conditions of incomplete input variables.
\begin{figure} 
	\centering
	\includegraphics[width=9cm]{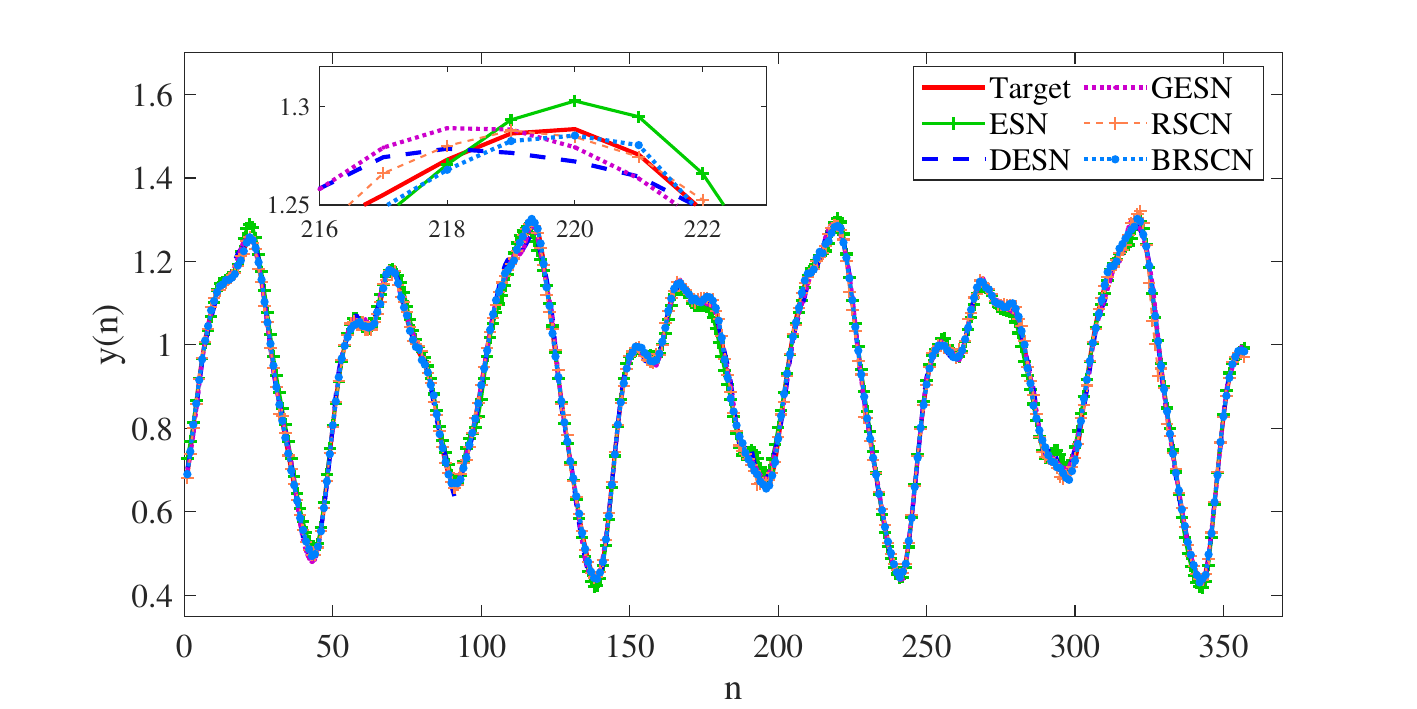}
	\caption{Prediction fitting curves of each model on MG2 task.}
	\label{fig4}
 \vspace{-0.5cm}
\end{figure}

The size of the reservoir significantly affects the performance of the model. In our experiments, we analyze the validation NRMSE curves of different models to identify the optimal reservoir size. Furthermore, the performance of BRSCNs with varying subreservoir sizes is also considered. Fig. \ref{fig3} displays the validation performance of various models on the MG task. It can be seen that the validation NRMSE decreases gradually as the reservoir sizes increase, suggesting underfitting. However, with an increase in the number of nodes, the validation NRMSE surpasses the minimum point, indicating overfitting. The validation performance of BRSCN varies with different subreservoir sizes, and the best result can be obtained when the subreservoir size is 10. Evidently, the optimal reservoir sizes for the ESN, DESN, GESN, RSCN, and BRSCN are determined to be 96, 80, 90, 68, and 60, respectively. Moreover, the validation performance of the BRSCN always outperforms other models, highlighting that the proposed method can contribute to sound performance if certain tricks are adopted to prevent overfitting.
\begin{figure*} 
	\centering
	\subfloat[MG]{\includegraphics[width=5.5cm]{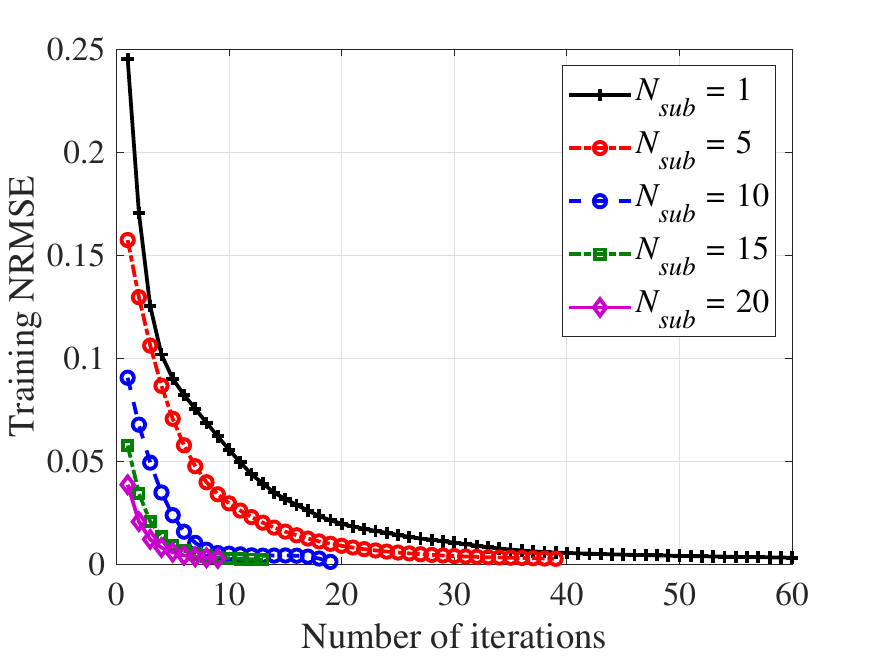}}
	\subfloat[MG1]{\includegraphics[width=5.5cm]{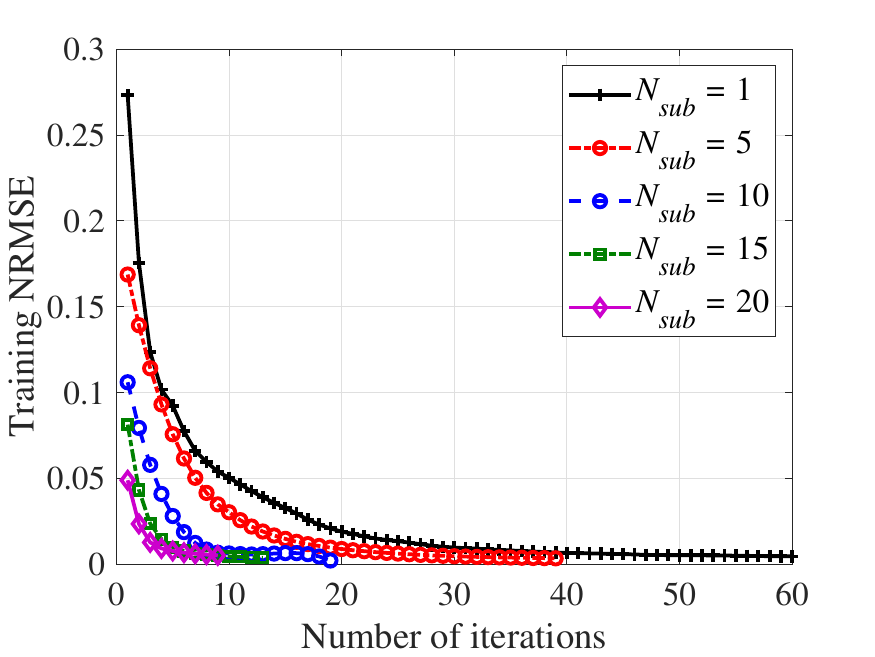}}
        \subfloat[MG2]{\includegraphics[width=5.5cm]{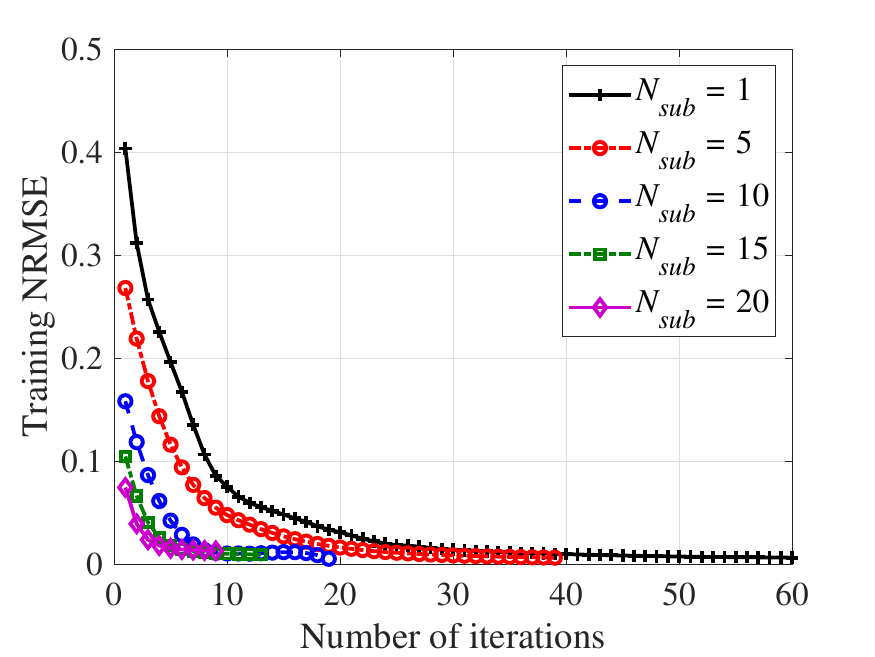}}
	\caption{Convergence performance of BRSCN with different subreservoir sizes on MG tasks.}
	\label{fig5}
 \vspace{-0.3cm}
\end{figure*}

Fig. \ref{fig4} illustrates the prediction fitting curves of each model for the MG2 task. It is clear that the BRSCN shows the highest degree of fitting with the target. This observation suggests that the proposed method has superior predictive capabilities compared to other models. Fig.~\ref{fig5} depicts the error convergence performance of BRSCNs with different subreservoir sizes on the three MG tasks. The model with ${N_{{\rm{sub}}}} = 1$ represents the original RSCN, where ${N_{{\rm{sub}}}}$ is the subreservoir size. We can see that a larger ${N_{{\rm{sub}}}}$ results in fewer iterations and higher rates of residual error reduction, suggesting that the block incremental construction can effectively improve the modelling efficiency.

Table~\ref{tb1} presents a comprehensive performance comparison of various models across the three MG tasks. The NRMSE of all models shows a gradual increase in the MG-MG2 tasks, indicating that each significant input variable contributes to the final output. The BRSCN stands out with a smaller reservoir size and lower training and testing NRMSE (except for the MG2 task) compared to other methods. These findings demonstrate the ability of BRSCNs to construct a more compact reservoir and reduce the impact of order uncertainty. Additionally, BRSCNs offer notable benefits in terms of modelling flexibility and efficiency, enabling various sub-modules to handle different features independently and streamline the calculation process. This ultimately reduces the utilization of computing resources, confirming their efficacy in time series forecasting.

\subsection{Nonlinear system identification}
In this simulation, the nonlinear system identification is considered. The dynamic nonlinear plant can be expressed by
\begin{equation} \label{eq30}
\begin{array}{l}
y\left( {n + 1} \right) = 0.72y\left( n \right) + 0.025y\left( {n - 1} \right)u\left( {n - 1} \right)\\
{\kern 1pt} {\kern 1pt} {\kern 1pt} {\kern 1pt} {\kern 1pt} {\kern 1pt} {\kern 1pt} {\kern 1pt} {\kern 1pt} {\kern 1pt} {\kern 1pt} {\kern 1pt} {\kern 1pt} {\kern 1pt} {\kern 1pt} {\kern 1pt} {\kern 1pt} {\kern 1pt} {\kern 1pt} {\kern 1pt} {\kern 1pt} {\kern 1pt} {\kern 1pt} {\kern 1pt} {\kern 1pt} {\kern 1pt} {\kern 1pt} {\kern 1pt} {\kern 1pt} {\kern 1pt} {\kern 1pt} {\kern 1pt} {\kern 1pt} {\kern 1pt} {\kern 1pt} {\kern 1pt} {\kern 1pt} {\kern 1pt} {\kern 1pt} {\kern 1pt} {\kern 1pt} {\kern 1pt} {\kern 1pt} {\kern 1pt} {\kern 1pt} {\kern 1pt} {\kern 1pt} {\kern 1pt} {\kern 1pt} {\kern 1pt} {\kern 1pt} {\kern 1pt}  + 0.01{u^2}\left( {n - 2} \right) + 0.2u\left( {n - 3} \right).
\end{array}
\end{equation}
In the training phase, $u\left( n \right)$ is uniformly generated from $\left[ { - 1,1} \right]$, and the initial output values are set to: $y\left( 1 \right)=y\left( 2 \right)=y\left( 3 \right)=0, y\left( 4 \right)=0.1.$ In the testing phase, the input is generated by
\begin{equation} \label{eq31}
u\left( n \right) = \left\{ {\begin{array}{*{20}{l}}
{\sin \left( {\frac{{\pi n}}{{25}}} \right),{\kern 1pt} {\kern 1pt} {\kern 1pt} {\kern 1pt} {\kern 1pt} {\kern 1pt} {\kern 1pt} {\kern 1pt} {\kern 1pt} {\kern 1pt} {\kern 1pt} {\kern 1pt} {\kern 1pt} {\kern 1pt} {\kern 1pt} {\kern 1pt} {\kern 1pt} {\kern 1pt} {\kern 1pt} {\kern 1pt} {\kern 1pt} {\kern 1pt} {\kern 1pt} {\kern 1pt} {\kern 1pt} {\kern 1pt} {\kern 1pt} {\kern 1pt} {\kern 1pt} {\kern 1pt} {\kern 1pt} {\kern 1pt} {\kern 1pt} {\kern 1pt} {\kern 1pt} {\kern 1pt} {\kern 1pt} {\kern 1pt} {\kern 1pt} {\kern 1pt} {\kern 1pt} {\kern 1pt} {\kern 1pt} {\kern 1pt} {\kern 1pt} {\kern 1pt} {\kern 1pt} {\kern 1pt} {\kern 1pt} {\kern 1pt} {\kern 1pt} {\kern 1pt} {\kern 1pt} {\kern 1pt} {\kern 1pt} {\kern 1pt} {\kern 1pt} {\kern 1pt} {\kern 1pt} {\kern 1pt} {\kern 1pt} {\kern 1pt} {\kern 1pt} {\kern 1pt} {\kern 1pt} {\kern 1pt} {\kern 1pt} {\kern 1pt} {\kern 1pt} {\kern 1pt} {\kern 1pt} {\kern 1pt} {\kern 1pt} {\kern 1pt} {\kern 1pt} {\kern 1pt} {\kern 1pt} {\kern 1pt} {\kern 1pt} {\kern 1pt} {\kern 1pt} {\kern 1pt} {\kern 1pt} {\kern 1pt} {\kern 1pt} {\kern 1pt} {\kern 1pt} {\kern 1pt} {\kern 1pt} {\kern 1pt} {\kern 1pt} {\kern 1pt} {\kern 1pt} {\kern 1pt} {\kern 1pt} {\kern 1pt} {\kern 1pt} {\kern 1pt} {\kern 1pt} {\kern 1pt} {\kern 1pt} {\kern 1pt} 0 < n < 250}\\
{1,{\kern 1pt}  {\kern 1pt} {\kern 1pt} {\kern 1pt} {\kern 1pt} {\kern 1pt} {\kern 1pt} {\kern 1pt} {\kern 1pt} {\kern 1pt} {\kern 1pt} {\kern 1pt} {\kern 1pt} {\kern 1pt} {\kern 1pt} {\kern 1pt} {\kern 1pt} {\kern 1pt} {\kern 1pt} {\kern 1pt} {\kern 1pt} {\kern 1pt} {\kern 1pt} {\kern 1pt} {\kern 1pt} {\kern 1pt} {\kern 1pt} {\kern 1pt} {\kern 1pt} {\kern 1pt} {\kern 1pt} {\kern 1pt} {\kern 1pt} {\kern 1pt} {\kern 1pt} {\kern 1pt} {\kern 1pt} {\kern 1pt} {\kern 1pt} {\kern 1pt} {\kern 1pt} {\kern 1pt} {\kern 1pt} {\kern 1pt} {\kern 1pt} {\kern 1pt} {\kern 1pt} {\kern 1pt} {\kern 1pt} {\kern 1pt} {\kern 1pt} {\kern 1pt} {\kern 1pt} {\kern 1pt} {\kern 1pt} {\kern 1pt}{\kern 1pt} {\kern 1pt} {\kern 1pt} {\kern 1pt} {\kern 1pt} {\kern 1pt} {\kern 1pt} {\kern 1pt} {\kern 1pt} {\kern 1pt} {\kern 1pt} {\kern 1pt} {\kern 1pt} {\kern 1pt} {\kern 1pt} {\kern 1pt} {\kern 1pt} {\kern 1pt} {\kern 1pt} {\kern 1pt} {\kern 1pt} {\kern 1pt} {\kern 1pt} {\kern 1pt} {\kern 1pt} {\kern 1pt} {\kern 1pt} {\kern 1pt} {\kern 1pt} {\kern 1pt} {\kern 1pt} {\kern 1pt} {\kern 1pt} {\kern 1pt} {\kern 1pt} {\kern 1pt} {\kern 1pt} {\kern 1pt} {\kern 1pt} {\kern 1pt} {\kern 1pt} {\kern 1pt} {\kern 1pt} {\kern 1pt} {\kern 1pt} {\kern 1pt} {\kern 1pt} {\kern 1pt} {\kern 1pt} {\kern 1pt} {\kern 1pt} {\kern 1pt} {\kern 1pt} {\kern 1pt} {\kern 1pt} {\kern 1pt} {\kern 1pt} {\kern 1pt} {\kern 1pt} {\kern 1pt} {\kern 1pt} {\kern 1pt} {\kern 1pt} {\kern 1pt} {\kern 1pt} {\kern 1pt} {\kern 1pt} {\kern 1pt} {\kern 1pt} {\kern 1pt} 250 \le n < 500}\\
{ - 1,{\kern 1pt}  {\kern 1pt} {\kern 1pt}  {\kern 1pt} {\kern 1pt} {\kern 1pt} {\kern 1pt} {\kern 1pt} {\kern 1pt} {\kern 1pt}{\kern 1pt} {\kern 1pt} {\kern 1pt} {\kern 1pt} {\kern 1pt} {\kern 1pt} {\kern 1pt} {\kern 1pt} {\kern 1pt} {\kern 1pt} {\kern 1pt} {\kern 1pt} {\kern 1pt} {\kern 1pt} {\kern 1pt} {\kern 1pt} {\kern 1pt} {\kern 1pt} {\kern 1pt} {\kern 1pt} {\kern 1pt} {\kern 1pt} {\kern 1pt} {\kern 1pt} {\kern 1pt} {\kern 1pt} {\kern 1pt} {\kern 1pt} {\kern 1pt} {\kern 1pt} {\kern 1pt} {\kern 1pt} {\kern 1pt} {\kern 1pt} {\kern 1pt} {\kern 1pt} {\kern 1pt} {\kern 1pt} {\kern 1pt} {\kern 1pt} {\kern 1pt} {\kern 1pt} {\kern 1pt} {\kern 1pt} {\kern 1pt} {\kern 1pt} {\kern 1pt} {\kern 1pt} {\kern 1pt} {\kern 1pt} {\kern 1pt} {\kern 1pt} {\kern 1pt} {\kern 1pt} {\kern 1pt} {\kern 1pt} {\kern 1pt} {\kern 1pt} {\kern 1pt} {\kern 1pt} {\kern 1pt} {\kern 1pt} {\kern 1pt} {\kern 1pt} {\kern 1pt} {\kern 1pt} {\kern 1pt} {\kern 1pt} {\kern 1pt} {\kern 1pt} {\kern 1pt} {\kern 1pt} {\kern 1pt} {\kern 1pt} {\kern 1pt} {\kern 1pt} {\kern 1pt} {\kern 1pt} {\kern 1pt} {\kern 1pt} {\kern 1pt} {\kern 1pt} {\kern 1pt} {\kern 1pt} {\kern 1pt} {\kern 1pt} {\kern 1pt} {\kern 1pt} {\kern 1pt} {\kern 1pt} {\kern 1pt} {\kern 1pt} {\kern 1pt} {\kern 1pt} {\kern 1pt} {\kern 1pt} {\kern 1pt} {\kern 1pt} {\kern 1pt} {\kern 1pt} {\kern 1pt} {\kern 1pt} {\kern 1pt} {\kern 1pt} {\kern 1pt} {\kern 1pt} {\kern 1pt} {\kern 1pt} 500 \le n < 750}\\
\begin{array}{l}
0.6\cos \left( {\frac{{\pi n}}{{10}}} \right) + 0.1\cos \left( {\frac{{\pi n}}{{32}}} \right) + \\
{\kern 1pt} {\kern 1pt} {\kern 1pt} {\kern 1pt} {\kern 1pt} 0.3\sin \left( {\frac{{\pi n}}{{25}}} \right),{\kern 1pt} {\kern 1pt} {\kern 1pt} {\kern 1pt} {\kern 1pt} {\kern 1pt} {\kern 1pt} {\kern 1pt} {\kern 1pt} {\kern 1pt} {\kern 1pt} {\kern 1pt} {\kern 1pt}  {\kern 1pt} {\kern 1pt} {\kern 1pt} {\kern 1pt} {\kern 1pt} {\kern 1pt} {\kern 1pt} {\kern 1pt} {\kern 1pt} {\kern 1pt} {\kern 1pt} {\kern 1pt} {\kern 1pt} {\kern 1pt} {\kern 1pt} {\kern 1pt} {\kern 1pt} {\kern 1pt} {\kern 1pt} {\kern 1pt} {\kern 1pt} {\kern 1pt} {\kern 1pt} {\kern 1pt} {\kern 1pt} {\kern 1pt} {\kern 1pt} {\kern 1pt} {\kern 1pt} {\kern 1pt} {\kern 1pt} {\kern 1pt} {\kern 1pt} {\kern 1pt} {\kern 1pt} {\kern 1pt} {\kern 1pt} {\kern 1pt} {\kern 1pt} {\kern 1pt} {\kern 1pt} {\kern 1pt} {\kern 1pt} {\kern 1pt} {\kern 1pt} {\kern 1pt} {\kern 1pt} {\kern 1pt} {\kern 1pt} {\kern 1pt} {\kern 1pt} {\kern 1pt} {\kern 1pt} {\kern 1pt} 750 \le n \le 1000.
\end{array}
\end{array}} \right.
\end{equation}
Considering the order uncertainty, only $y\left( n \right)$ and $u\left( n \right)$ are used to predict $y\left( n+1 \right)$. The training, validation, and testing set consists of 2000, 1000, and 1000 samples, respectively. The first 100 samples of each set are washed out.
\begin{figure} 
	\centering
         \includegraphics[width=9cm]{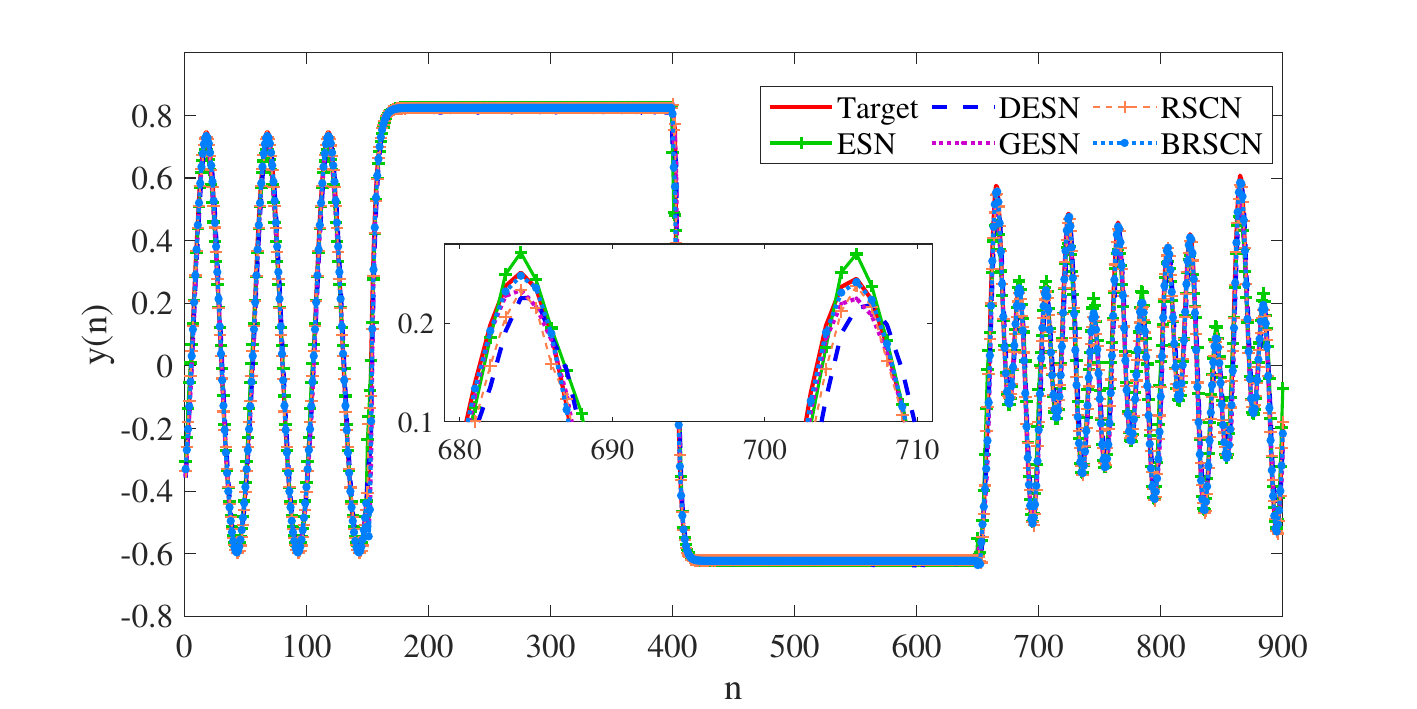}
	\caption{The prediction curves of each model for the nonlinear system identification.}
	\label{fig6}
 \vspace{-0.5cm}
\end{figure}
\begin{figure} 
	\centering
	\includegraphics[width=5.5cm]{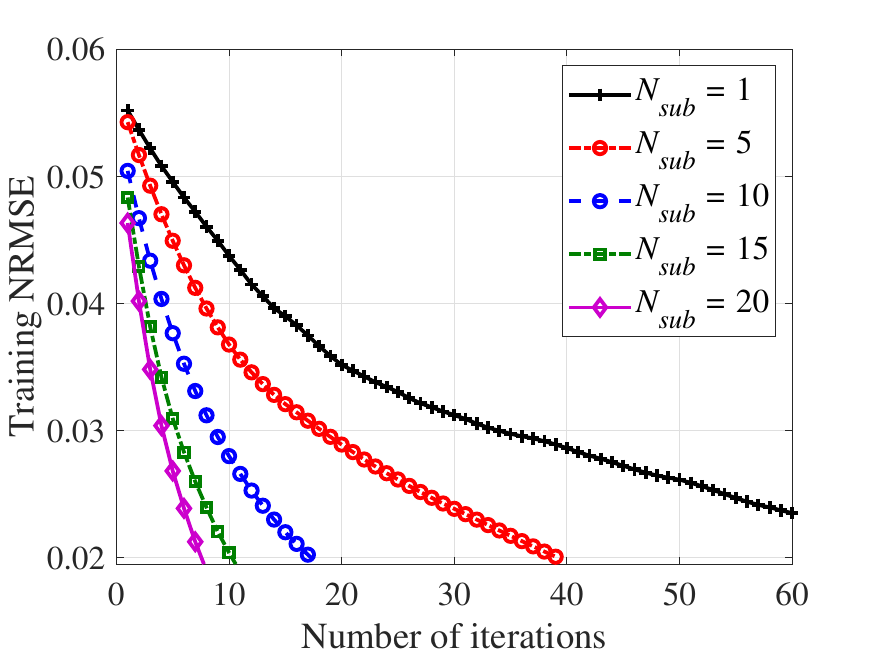}
	\caption{Convergence performance of BRSCN with different subreservoir sizes on the nonlinear system identification task.}
	\label{fig8}
 \vspace{-0.8cm}
\end{figure}

The prediction curves for each model on the nonlinear system identification task are shown in Fig.~\ref{fig6}. The predicted output of BRSCNs have a higher fitting degree to the desired output than other models. These results suggest that BRSCNs can quickly respond to the dynamic variations in the system and are well-suited for modelling nonlinear dynamic systems. Fig.~\ref{fig8} illustrates the error convergence performance of BRSCN with various subreservoir sizes. Compared with the original RSCN (BRSCN with ${N_{{\rm{sub}}}} = 1$), BRSCNs demonstrate higher residual error decreasing rates and require fewer iterations, thus confirming the superior modelling efficiency of the proposed method.

To comprehensively compare and analyze the modelling performance of the proposed methods for the two nonlinear system identification tasks, the experimental results of various models are listed in Table~\ref{tb2}. Obviously, our proposed BRSCNs outperform other models in terms of training and testing NRMSE, and BRSCNs with ${N_{{\rm{sub}}}} = 10$ exhibit the best performance. Specifically, the training and testing NRMSE obtained by BRSCN account for 17.03$\%$ and 53.51$\%$ of that obtained by the ESN, and 19.65$\%$ and 84.84$\%$ of that obtained by the RSCN. Moreover, BRSCNs can achieve the highest prediction accuracy with the smallest reservoir sizes, resulting in a more compact topology. These results underscore the significant potential of BRSCNs in identifying nonlinear systems, particularly those with unknown dynamic orders.

\begin{table*}[htbp]
\caption{Performance comparison of different models on the nonlinear system identification.} \label{tb2}
\centering
\begin{tabular}{ccccc}
\hline
Models & Reservoir size & Training time            & Training NRMSE           & Testing NRMSE            \\ \hline
ESN    & 157            & 0.06339±0.01033          & 0.00916±0.00252          & 0.06276±0.00325          \\
DESN   & 110            & \textbf{0.05237±0.00938}          & 0.00818±0.00062          & 0.05493±0.00622          \\
GESN   & 120            & 1.34652±0.12631 & 0.00805±0.00049          & 0.04316±0.00128          \\
RSCN   & 102            & 2.23162±0.85353          & 0.00794±0.00057          & 0.03958±0.00097          \\
BRSCN (${N_{{\rm{sub}}}} = 5$)  & 105            & 2.00807±0.82080          & 0.00217±0.00031 & 0.03569±0.00695 \\
BRSCN (${N_{{\rm{sub}}}} = 10$) & 110            & 1.88672±0.50024          & \textbf{0.00156±0.00021}          & \textbf{0.03358±0.00377}          \\
BRSCN (${N_{{\rm{sub}}}} = 15$)  & 90             & 1.52153±0.48446          & 0.00221±0.00043          & 0.03526±0.00496          \\
BRSCN (${N_{{\rm{sub}}}} = 20$) & 100            & 1.53395±0.63698          & 0.00159±0.00029          & 0.03798±0.00362          \\ \hline
\end{tabular}
\vspace{-0.5cm}
\end{table*}

\subsection{Soft sensing of the butane concentration in the debutanizer column process}
\begin{table}[ht]
 \centering
  \caption{Process auxiliary variables for the debutanizer column.}
  \label{tb3}
\begin{tabular}{cc}
\hline
Input variables & Variable description      \\ \hline
${{u_1}}$             & tower top   temperature   \\
${{u_2}}$              & tower top   pressure      \\
${{u_3}}$             & tower top   reflux flow   \\
${{u_4}}$             & tower top product outflow \\
${{u_5}}$             & 6-th tray   temperature   \\
${{u_6}}$            & tower bottom temperature  \\
${{u_7}}$             & tower bottom pressure     \\ \hline
\end{tabular}
\end{table}
\begin{figure}[htbp]
	\begin{center}
		\includegraphics[width=6cm]{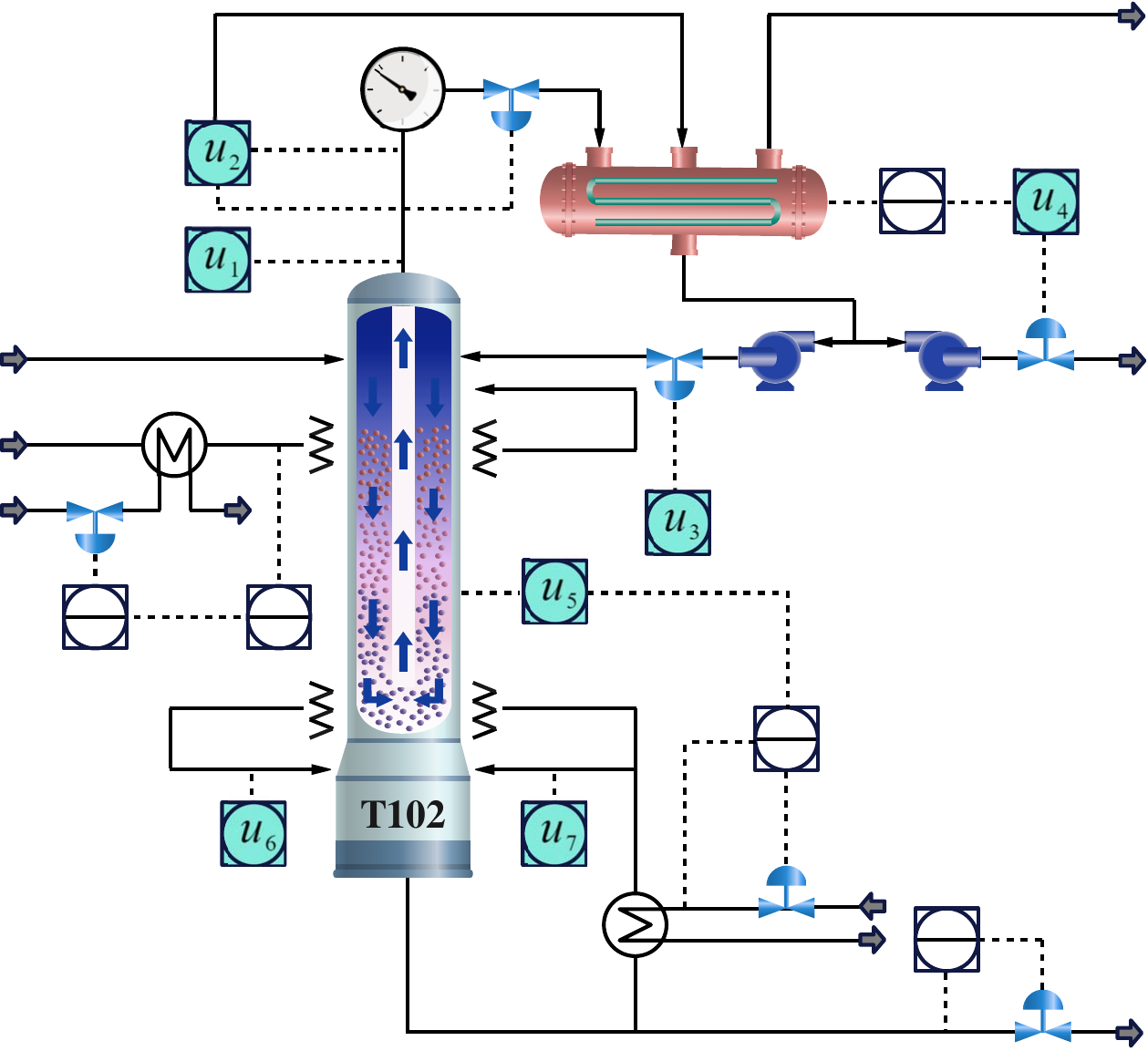}
		\caption{Flowchart of debutanizer column process.}
		\label{fig9}
	\end{center}
 \vspace{-0.5cm}
\end{figure}
\begin{figure} 
	\centering
         \includegraphics[width=9cm]{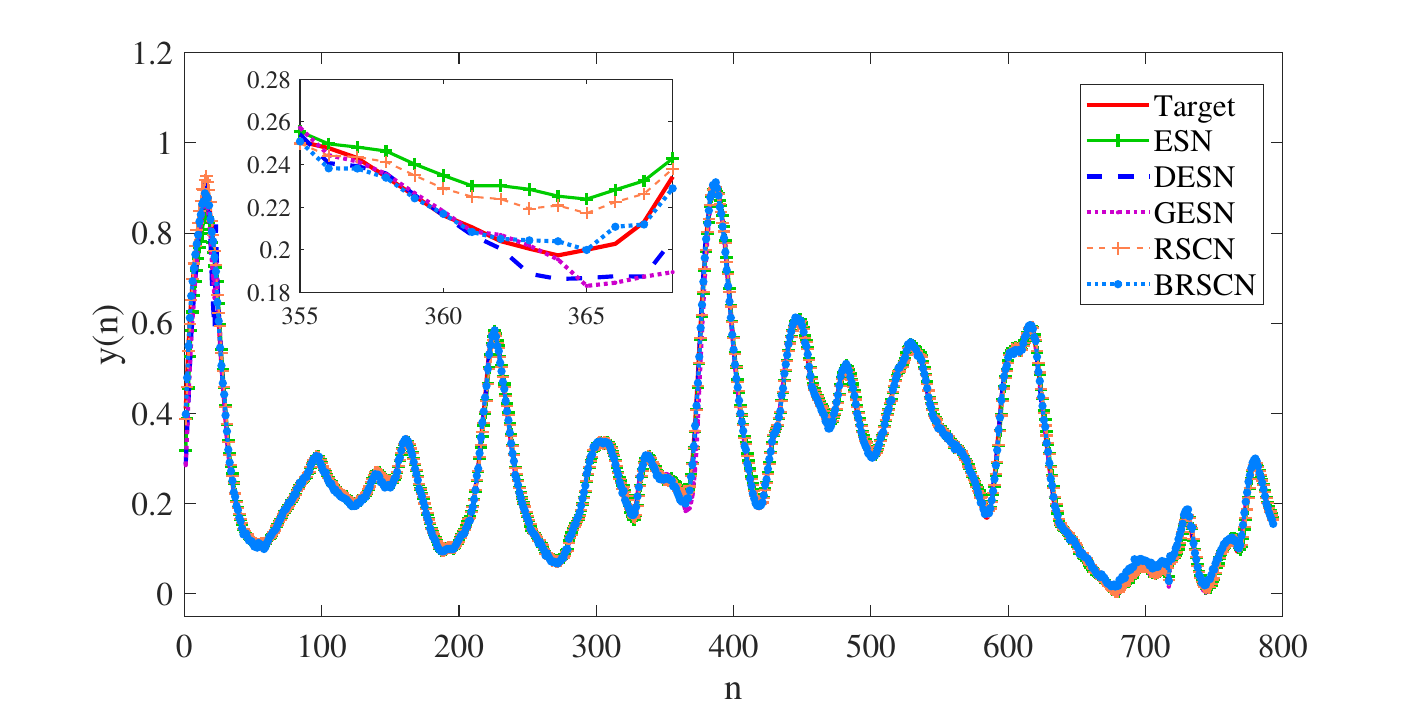}
	\caption{The prediction curves of each model for debutanizer column process.}
	\label{fig11}
 \vspace{-0.5cm}
\end{figure}
The debutanizer column is an important refining unit in petrochemical plants, aiming at splitting naphtha and desulfurizing. However, fluctuations in the external environment and changes in raw material composition can lead to variations in direct measurements, impacting the control and improvement of the production process. Thus, it is crucial to establish a precise soft sensing model. As shown in Fig.~\ref{fig9}, this process mainly includes six devices: overhead condenser, heat exchanger, tower top reflux pump, bottom reboiler, reflux accumulator, and feed pump liquefied petroleum gas separator. Table~\ref{tb3} lists the seven relevant process variables for modelling and predicting C4 content during the production process. In \cite{ref25}, Fortuna et al. introduced a well-designed combination of variables to obtain the butane concentration $y$,\vspace{-0.35cm}

\begin{small}
    \begin{equation} \label{eq37}
\begin{array}{l}
y\left( n \right) = f\left( {{u_1}\left( n \right),} \right.{u_2}\left( n \right),{u_3}\left( n \right),{u_4}\left( n \right),{u_5}\left( n \right),{u_5}\left( {n - 1} \right),\\
{\kern 1pt} {\kern 1pt} {\kern 1pt} {\kern 1pt} {\kern 1pt} {\kern 1pt} {\kern 1pt} {\kern 1pt} {\kern 1pt} {\kern 1pt} {\kern 1pt} {\kern 1pt} {\kern 1pt} {\kern 1pt} {\kern 1pt} {\kern 1pt} {\kern 1pt} {\kern 1pt} {\kern 1pt} {\kern 1pt} {\kern 1pt} {\kern 1pt} {\kern 1pt} {\kern 1pt} {\kern 1pt} {\kern 1pt} {\kern 1pt} {\kern 1pt} {\kern 1pt} {\kern 1pt} {\kern 1pt} {\kern 1pt} {\kern 1pt} {\kern 1pt} {\kern 1pt} {\kern 1pt} {\kern 1pt} {\kern 1pt} {\kern 1pt} {\kern 1pt} {\kern 1pt} {\kern 1pt} {\kern 1pt} {\kern 1pt} {\kern 1pt} {\kern 1pt} {\kern 1pt} {\kern 1pt} {\kern 1pt} {\kern 1pt} {u_5}\left( {n - 2} \right),{u_5}\left( {n - 3} \right),\left( {{u_1}\left( n \right) + {u_2}\left( n \right)} \right)/2,\\
{\kern 1pt} {\kern 1pt} {\kern 1pt} {\kern 1pt} {\kern 1pt} {\kern 1pt} {\kern 1pt} {\kern 1pt} {\kern 1pt} {\kern 1pt} {\kern 1pt} {\kern 1pt} {\kern 1pt} {\kern 1pt} {\kern 1pt} {\kern 1pt} {\kern 1pt} {\kern 1pt} {\kern 1pt} {\kern 1pt} {\kern 1pt} {\kern 1pt} {\kern 1pt} {\kern 1pt} {\kern 1pt} {\kern 1pt} {\kern 1pt} {\kern 1pt} {\kern 1pt} {\kern 1pt} {\kern 1pt} {\kern 1pt} {\kern 1pt} {\kern 1pt} {\kern 1pt} {\kern 1pt} {\kern 1pt} {\kern 1pt} {\kern 1pt} {\kern 1pt} {\kern 1pt} {\kern 1pt} {\kern 1pt} {\kern 1pt} {\kern 1pt} {\kern 1pt} {\kern 1pt} \left. {{\kern 1pt} {\kern 1pt} {\kern 1pt} y\left( {n - 1} \right),y\left( {n - 2} \right),y\left( {n - 3} \right),y\left( {n - 4} \right)} \right).
\end{array}
\end{equation}
\end{small}In our experiments, the order uncertainty is considered and the input is set as $\left[ {{u}_{1}}\left( n \right), \right.{{u}_{2}}\left( n \right),{{u}_{3}}\left( n \right),{{u}_{4}}\left( n \right), {{u}_{5}}\left( n \right),$ $\left.y\left( n-1 \right) \right]$. The dataset consists of 2394 samples, divided into the first 1500 samples for training and the last 894 samples for testing. The Gaussian white noise is added to the testing set to generate the validation set. The first 100 samples of each set are washed out.

Fig.~\ref{fig11} shows the prediction curves of each model for the debutanizer column process. It can be seen that the BRSCN achieves a higher degree of accuracy in approximating the desired output compared to the other models. This improvement underscores the effectiveness of the block incremental strategy, highlighting its potential in managing the complexities and nonlinearities inherent in the real industry processes. 

\subsection{Short-term power load forecasting}
\begin{figure}[htbp]
	\begin{center}
		\includegraphics[width=7cm]{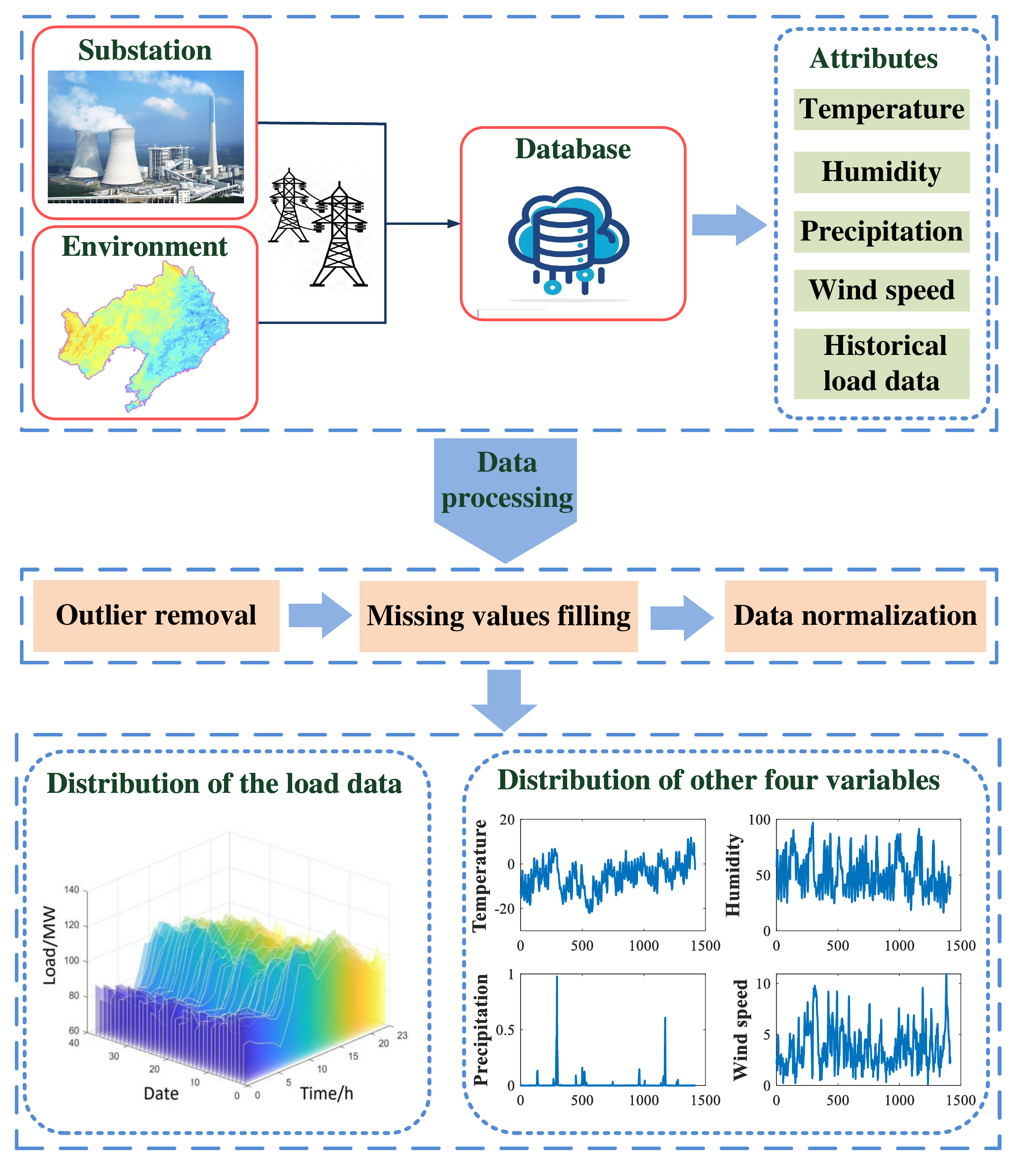}
		\caption{Flowchart of the data collection and processing for the short-term power load forecasting.}
		\label{fig10}
	\end{center}
    \vspace{-0.3cm}
\end{figure}
Short-term power load forecasting is crucial for the reliable and efficient operation of power systems and cost reduction. This study analyzes load data from a 500kV substation in Liaoning Province, China, which is collected hourly from January to February 2023, spanning a total of 59 days. Environmental factors like temperature ${u_1}$, humidity ${u_2}$, precipitation ${u_3}$, and wind speed ${u_4}$ are considered to forecast power load $y$. The process of data collection and analysis for short-term electricity load forecasting is illustrated in Fig.~\ref{fig10}. The dataset consists of 1415 samples, with 1000 for training and 415 for testing. Gaussian noise is introduced to the testing set to create the validation set. Considering order uncertainty, $\left[ {{u}_{1}}\left( n \right), \right.$$\left. {{u}_{2}}\left( n \right),{{u}_{3}}\left( n \right),{{u}_{4}}\left( n \right),y\left( n-1 \right) \right]$ is utilized to predict $y\left( n \right)$ in the experiment. The initial 30 samples from each set are disregarded.

\begin{figure}
	\centering
         \includegraphics[width=9cm]{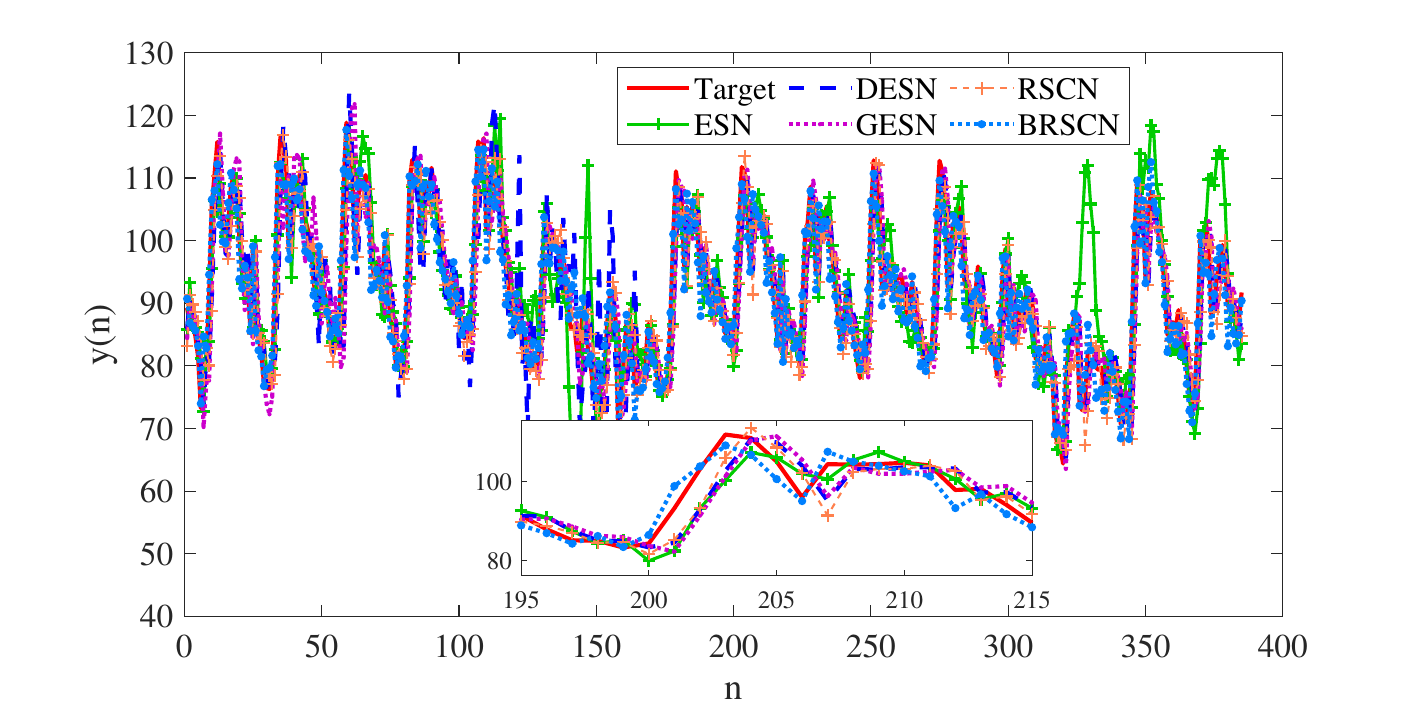}
	\caption{The prediction curves of each model for the short-term power load forecasting.}
	\label{fig12}
\vspace{-0.5cm}
\end{figure}

\begin{figure}[htbp]
\vspace{-0.3cm}
\centering
	\subfloat[Case 1]{\includegraphics[width=5.5cm]{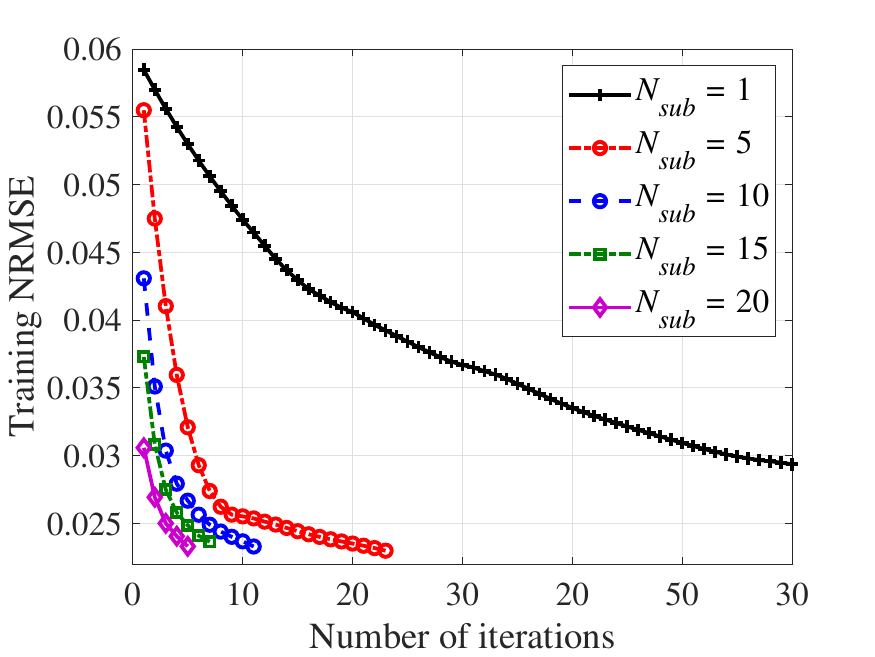}}\\ \vspace{-0.2cm}
	\subfloat[Case 2]{\includegraphics[width=5.5cm]{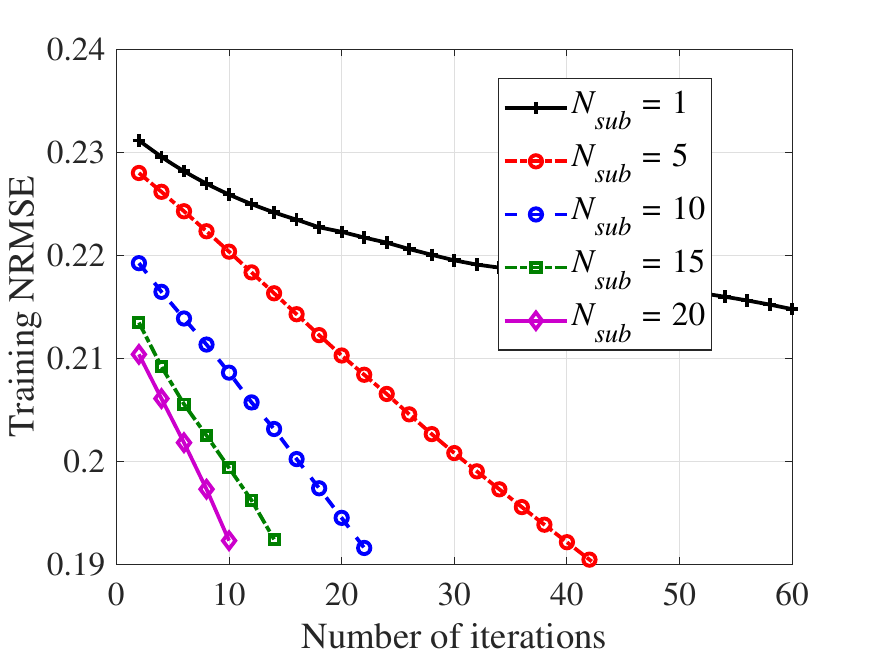}}
	\caption{Convergence performance of BRSCN with different subreservoir sizes on the two industry cases.}
\label{fig13}
\vspace{-0.2cm}
\end{figure}

Fig.~\ref{fig12} exhibits the prediction curves of various models on the short-term power load forecasting. It is easy to see that compared with other models, the output of the BRSCN can better fit the desired output, thereby proving the feasibility of the proposed method for forecasting the industrial process parameters. Fig.~\ref{fig13} displays the error convergence performance of BRSCNs with varying subreservoir sizes on the different industry cases, demonstrating that larger block sizes result in fewer iterations and faster convergence speed.

To intuitively compare the modelling performance of the proposed BRSCNs with other models, we summarize the experimental results in Table~\ref{tb4}. It can be seen that BRSCNs exhibit the smallest NRMSE in both training and testing. Specifically, compared with the original RSCN, the reservoir topology of BRSCN is more compact, and the computational cost is also smaller. These results demonstrate the effectiveness and efficiency of the proposed BRSCNs in handling the industry process data, indicating their great potential for tackling complex dynamic modelling problems in practical industrial applications.
\begin{table*}[htbp]
\caption{Performance comparison of different models on debutanizer column process.} \label{tb4}
\centering
\begin{tabular}{cccccc}
\hline
Datasets                & Models & Reservoir size & Training time            & Training NRMSE           & Testing NRMSE            \\ \hline
\multirow{8}{*}{Case 1} & ESN    & 213            & 0.40819±0.12035          & 0.04085±0.00137          & 0.08427±0.01144          \\
                        & DESN    & 190            & \textbf{0.29232±0.03847} & 0.03063±0.00098          & 0.07193±0.01029          \\
                        & GESN   & 180            & 0.84227±0.19173          & 0.02905±0.00096          & 0.07060±0.00930          \\
                        & RSCN   & 87             & 2.23734±0.20122          & 0.02734±0.00100          & 0.06793±0.00411          \\
                        & BRSCN (${N_{{\rm{sub}}}} = 5$)  & 95             & 2.16321±0.23011          & 0.02566±0.00092          & 0.05726±0.00385          \\
                        & BRSCN (${N_{{\rm{sub}}}} = 10$)  & 80             & 2.18279±0.17355          & \textbf{0.02387±0.00088} & 0.05392±0.00382          \\
                        & BRSCN (${N_{{\rm{sub}}}} = 15$)  & 75             & 2.12527±0.18992          & 0.02412±0.00022          & \textbf{0.05283±0.00293} \\
                        & BRSCN (${N_{{\rm{sub}}}} = 20$)  & 80             & 2.12836±0.16359          & 0.02436±0.00025          & 0.05429±0.00283          \\ \hline
\multirow{8}{*}{Case 2} & ESN    & 103            & 0.11881±0.01548          & 0.24338±0.02189          & 0.64522±0.07172          \\
                        & DESN    & 90             & \textbf{0.08563±0.00945} & 0.21997±0.01398          & 0.58930±0.04126          \\
                        & GESN   & 110            & 0.56325±0.09371          & 0.22083±0.02835          & 0.60361±0.03328          \\
                        & RSCN   & 65             & 0.68257±0.03141          & 0.19254±0.02047          & 0.51078±0.03685          \\
                        & BRSCN (${N_{{\rm{sub}}}} = 5$)  & 55             & 0.63821±0.02790          & 0.19128±0.02163          & 0.49273±0.02236          \\
                        & BRSCN (${N_{{\rm{sub}}}} = 10$)  & 50             & 0.61289±0.02938          & \textbf{0.19088±0.02029} & \textbf{0.49099±0.01028} \\
                        & BRSCN (${N_{{\rm{sub}}}} = 15$)  & 60             & 0.61505±0.03271          & 0.19202±0.02739          & 0.49116±0.02837          \\
                        & BRSCN (${N_{{\rm{sub}}}} = 20$)  & 60             & 0.60398±0.01092          & 0.19391±0.03028          & 0.49325±0.02639          \\ \hline
\end{tabular}
\end{table*}

\begin{figure*} 
\vspace{-0.3cm}
	\centering
	\subfloat[Case1]{\includegraphics[width=7cm]{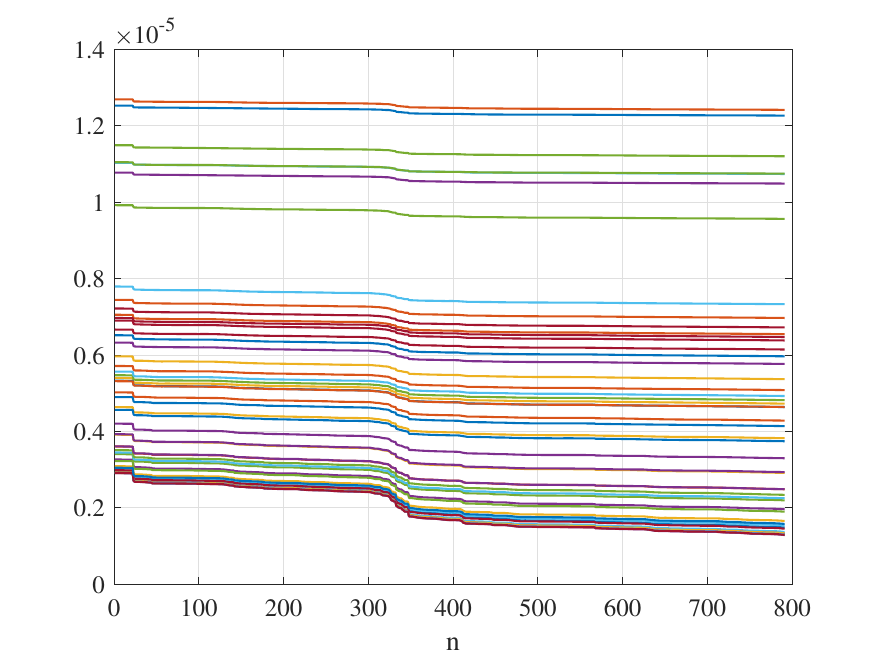}}
        \subfloat[Case2]{\includegraphics[width=7cm]{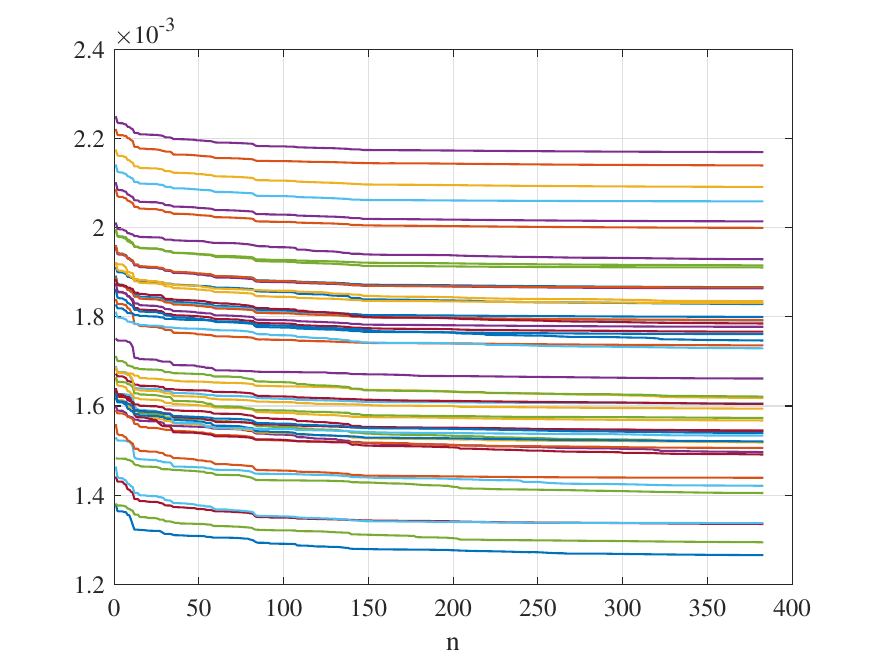}}
	\caption{Errors between the output weights updated by the projection algorithm
and trained offline on the two industry cases.}
	\label{fig15}
\vspace{-0.3cm}
\end{figure*}

Fig.~\ref{fig15} illustrates the errors in output weights updated by the projection algorithm compared to the weights trained offline for two industrial cases. We conducted 50 independent experiments, and the results clearly demonstrate the convergence of the output weights. This convergence is crucial for preventing oscillatory or unstable behavior in the model when deployed in practical industrial applications. Such a property is vital for ensuring the smooth operation of systems, particularly in industrial settings that require high precision and reliability. This online learning scheme facilitates a swift response to dynamic changes within the system, ensuring robust performance when handling complex industry data.

\section{Conclusion}
Efficient and accurate modelling is crucial for nonlinear dynamic systems with uncertain orders. This paper introduces an improved version of RSCN with block increments for problem solving. The proposed BRSCNs inherit the merits of the original RSCNs, such as data-dependent parameter selection, theoretical basis for structural design, and strong nonlinear approximation capabilities. From the implementation perspective, BRSCNs can simultaneously add multiple reservoir nodes in the light of a supervisory mechanism and each subreservoir is constructed with a special structure to guarantee the universal approximation property and echo state property of the built model. The output weights are updated online using the projection algorithm, and conditions for persistent excitation are outlined to guarantee parameter convergence. The effectiveness of the proposed approach is evaluated across four nonlinear dynamic modeling tasks. Experimental results demonstrate that BRSCNs can significantly enhance model compactness and achieve sound performance in terms of modelling efficiency, learning capability, and generalization.

This scheme will be applicable to address the data-driven modelling problems in the process industries, and it is interesting to see more real-world applications of the proposed technique. Furthermore, future research in this direction could investigate a regularized version of RSCNs or BRSCNs to enhance the model's ability to handle noisy and uncertain data.

\end{CJK}

\end{document}